\documentclass[runningheads]{llncs}

\usepackage{eccv}

\usepackage{eccvabbrv}

\usepackage{graphicx}
\usepackage{booktabs}
\usepackage{multirow}
\usepackage{array}
\usepackage{adjustbox}
\usepackage{pdflscape}
\usepackage{longtable}
\usepackage{float}
\usepackage{listings}
\usepackage{chngcntr}

\usepackage{xcolor}
\usepackage{pifont}

\usepackage[accsupp]{axessibility}

\usepackage{hyperref}
\hypersetup{hidelinks}

\usepackage{orcidlink}

\definecolor{listingbg}{RGB}{248,248,248}
\definecolor{listingframe}{RGB}{210,210,210}
\definecolor{listingnum}{RGB}{110,110,110}

\lstdefinestyle{promptstyle}{
  basicstyle=\ttfamily\scriptsize,
  backgroundcolor=\color{listingbg},
  frame=single,
  rulecolor=\color{listingframe},
  framerule=0.3pt,
  xleftmargin=0.8em,
  xrightmargin=0.8em,
  framesep=0.45em,
  breaklines=true,
  breakatwhitespace=false,
  columns=fullflexible,
  keepspaces=true,
  showstringspaces=false,
  tabsize=2,
  numbers=left,
  numberstyle=\tiny\color{listingnum},
  numbersep=6pt,
  numberblanklines=false,
  captionpos=t,
  aboveskip=0.8em,
  belowskip=0.4em
}

\newcommand{\equalcontrib}{\textsuperscript{\ensuremath{\dagger}}}
\newcommand{\linkicon}[1]{%
  \raisebox{-0.10em}{\includegraphics[height=0.78em]{#1}}%
}

\newcommand{\headerlink}[3]{%
  \href{#1}{\mbox{\linkicon{#2}\hspace{0.22em}#3}}%
}

\newcommand{\linksep}{%
  \hspace{0.45em}{\normalfont$\cdot$}\hspace{0.45em}%
}

\newcommand{\BeyondMasksLinks}{%
  {\small
  \headerlink{https://yigitekin.github.io/BeyondMasks/}{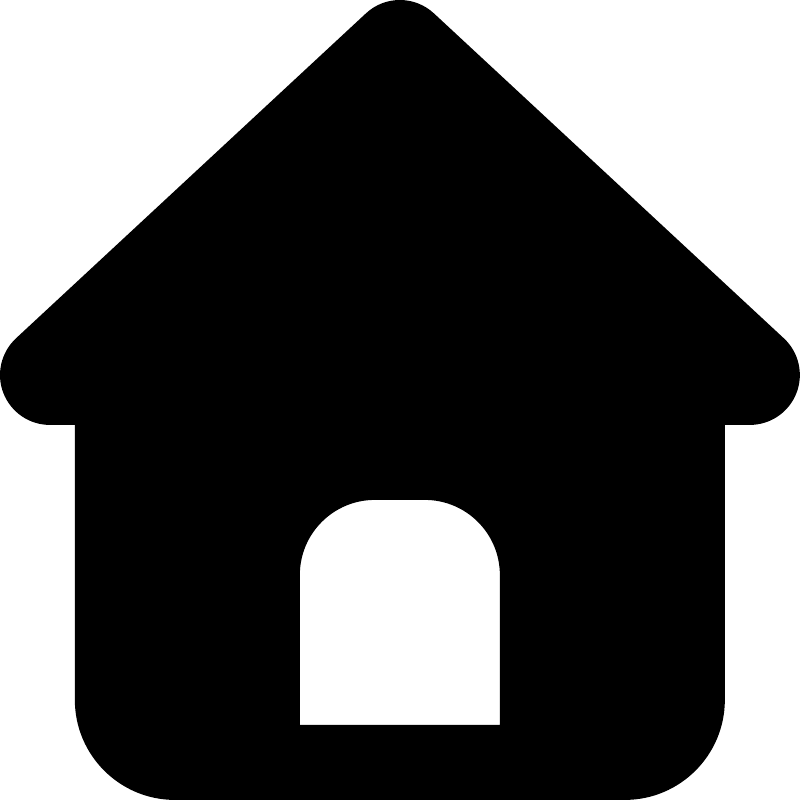}{Project Page}%
  \linksep
  \headerlink{https://github.com/YigitEkin/BeyondMasks}{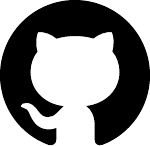}{Code}%
  \linksep
  \headerlink{https://huggingface.co/datasets/yigitekin/BeyondMasks/tree/main}{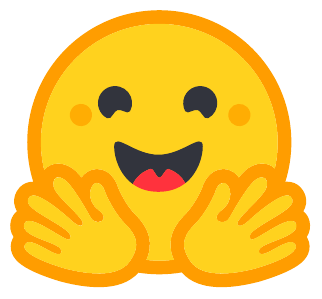}{Dataset}%
  }%
}

\newcommand{\Yes}{\textcolor{green!60!black}{\ding{51}}}
\newcommand{\No}{\textcolor{red!75!black}{\ding{55}}}

\begin{document}

\title{BeyondMasks: Evaluating Causal and Physical Consistency in Video Object Removal}

\titlerunning{BeyondMasks}

\author{
Yi\u{g}it Ekin\inst{1}\equalcontrib\orcidlink{0009-0006-4969-608X}
\and
Enes Sanli\inst{2,4}\equalcontrib\orcidlink{0009-0000-0361-6597}
\and
Aykut Erdem\inst{2,4}\orcidlink{0000-0002-6280-8422}
\and
Erkut Erdem\inst{3,4}\orcidlink{0000-0002-6744-8614}
\and
Aysegul Dundar\inst{1}\orcidlink{0000-0003-2014-6325}
}

\authorrunning{Y. Ekin et al.}

\institute{
Bilkent University, Department of Computer Engineering, Ankara, T\"urkiye
\and
Ko\c{c} University, Department of Computer Engineering, Istanbul, T\"urkiye
\and
Hacettepe University, Department of AI and Data Engineering, Ankara, T\"urkiye
\and
KUIS AI Center, Istanbul, T\"urkiye
\\
\email{yigit.ekin@bilkent.edu.tr} 
\\[0.35em]
\BeyondMasksLinks
}
\maketitle
\begingroup
\renewcommand{\thefootnote}{\ensuremath{\dagger}}
\footnotetext{Denotes equal contribution.}
\endgroup

\begin{figure}[H]
\centering
\includegraphics[width=\linewidth]{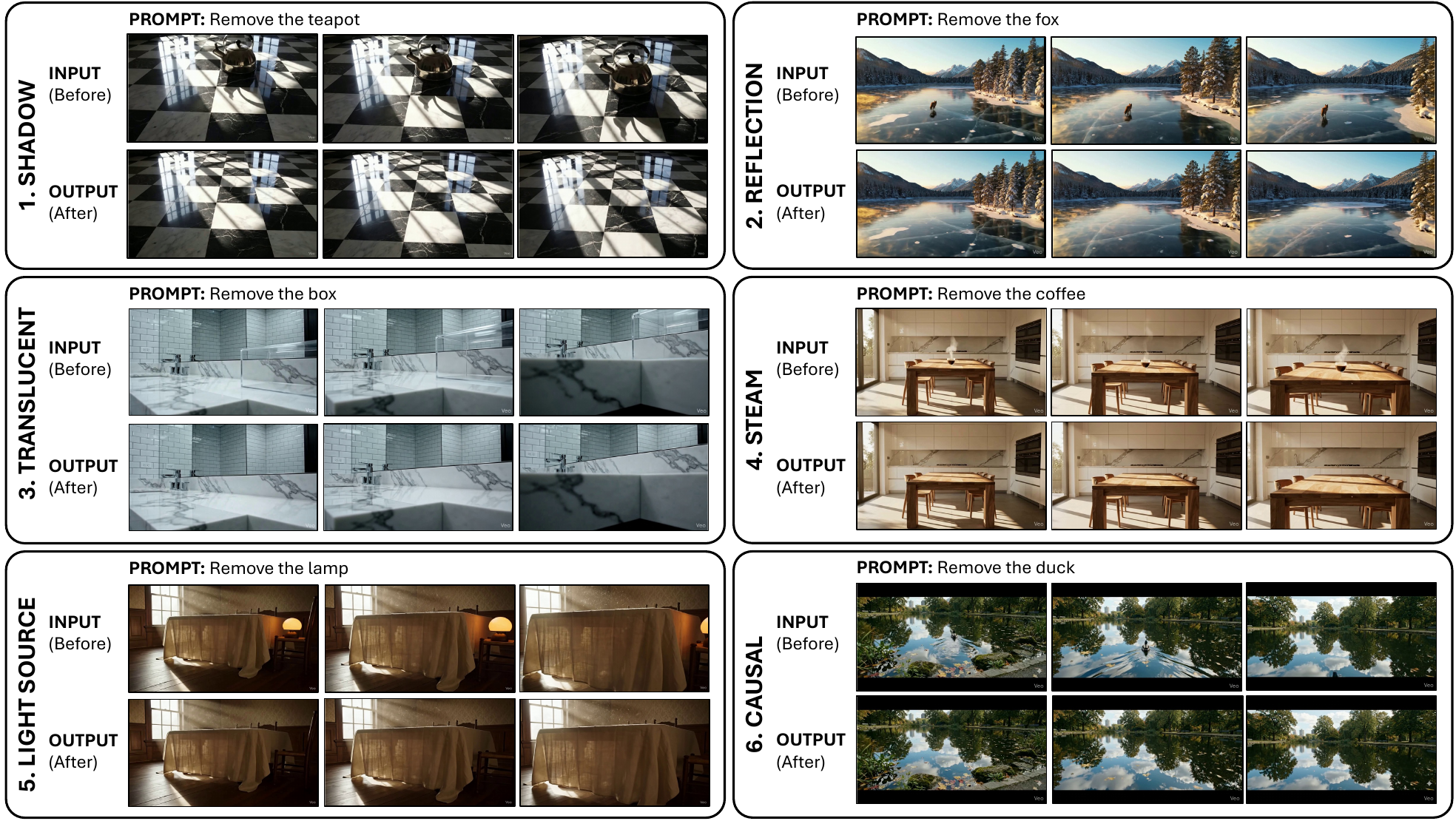}
\caption{\textbf{Video object removal requires eliminating both objects and their physical after-effects.} Objects interact with their environment through shadows, reflections, translucency, illumination changes, volumetric effects (e.g., steam), and dynamic physical traces. Removing the object alone does not restore the scene if these effects persist. BeyondMasks introduces a benchmark to evaluate such causal object-environment interactions using temporally aligned video pairs, enabling systematic assessment of whether methods recover the scene as if the object had never been present.
}
\label{fig:teaser}
\end{figure}

\begin{abstract}
Recent advances in generative video models have significantly improved visual realism in video object removal, yet evaluation protocols still focus on masked-region fidelity, treating removal as local inpainting. In real scenes, object removal is a causal intervention: eliminating an object also requires removing its induced physical effects, such as shadows, reflections, illumination changes, translucency, and dynamic traces. Existing benchmarks lack aligned clean references or remain limited to simplified synthetic settings, preventing systematic evaluation of causal consistency. We introduce \emph{BeyondMasks}, a paired benchmark for causally consistent video object removal, consisting of temporally aligned synthetic and real-world video pairs with clean background references. The dataset spans diverse photometric, geometric, volumetric, and dynamic interactions, and supports both mask-based and instruction-driven editing. We further propose \emph{CORE}, a structured vision–language model-based evaluation protocol that jointly measures object disappearance and after-effect consistency, aligning more closely with human judgments than existing metrics. Benchmarking state-of-the-art methods reveals systematic failures in removing secondary physical effects despite high masked-region fidelity, exposing a gap between visual plausibility and causal correctness. BeyondMasks reframes video object removal as causal scene consistency rather than local reconstruction and provides a unified framework for its evaluation.

\keywords{Video object removal with after effect\and Benchmark dataset \and Paired video evaluation \and Vision-language models}
\end{abstract}

\section{Introduction}
\label{sec:intro}
Video object removal aims to edit a video such that a selected object appears as if it were never present. With the rise of large-scale generative video models \cite{dharmarajan2025dream2flow, yu2025veggie, yang2024cogvideox, wang2025lavie}, recent systems achieve striking perceptual realism, enabling increasingly sophisticated scene manipulation. Modern video foundation models further integrate instruction-driven editing \cite{jiang2025vace} and open-world reasoning, positioning object removal \cite{miao2025rose} as a core capability within general-purpose video generation pipelines. However, evaluation protocols have not kept pace with model capabilities. Most existing benchmarks assess reconstruction fidelity only within the masked region, implicitly framing object removal as a localized inpainting problem.

In real scenes, object removal is fundamentally a \emph{causal intervention}.
Objects do not merely occupy pixels; they interact with their environment across
space and time. As illustrated in Fig.~\ref{fig:teaser}, they cast shadows, alter
illumination, appear in reflections, scatter light through translucent materials,
and induce dynamic traces such as footprints or dust displacements. Removing an
object therefore requires eliminating both the object itself and the downstream
physical effects it induces. Evaluating masked-region reconstruction alone
overlooks this broader causal footprint and fails to measure whether the scene
has been physically restored.

Current benchmarks are ill-suited for this purpose. Widely used video datasets such as DAVIS~\cite{Perazzi_2016_CVPR} and YouTube-VOS~\cite{Xu_2018_ECCV} lack paired clean backgrounds, preventing full removal verification. More recent benchmarks like ROSE-Bench~\cite{miao2025rose} introduce synthetic paired data to study specific physical side effects. While representing important progress, these datasets remain limited in interaction diversity and real-world variability, and typically focus on isolated photometric effects. They do not systematically evaluate dynamic object-environment interactions, nor do they naturally support instruction-driven editing pipelines increasingly used in modern video foundation models.

Motivated by these limitations, we advocate a causally grounded formulation of video object removal. Under this view, the task requires disentangling the target object from the scene and undoing its induced environmental changes consistently over time. This perspective shifts evaluation from local reconstruction quality toward object-environment consistency and temporal physical reasoning.
To enable this paradigm, we introduce \emph{BeyondMasks}, a diagnostically focused benchmark for evaluating causal and physical consistency in video object removal. BeyondMasks contains a collection of temporally aligned paired videos, including synthetic sequences generated through controlled editing and real-world tripod-captured scenes. Rather than maximizing scale, we prioritize precise causal control, physical diversity, and clean reference alignment, enabling systematic evaluation of object-environment disentanglement. Each sequence provides a clean reference background, per-frame object masks, and instruction prompts, supporting both mask-based and text-guided editing scenarios.

Beyond paired pixel-level evaluation, we propose CORE (Causal Object Removal Evaluation) score, a vision-language model (VLM)-based evaluation metric that jointly measures object disappearance and after-effect consistency. We validate this protocol through human studies and inter-model agreement analysis, demonstrating its reliability and alignment with perceptual judgments.

Extensive benchmarking of state-of-the-art mask-based and instruction-driven models reveals systematic failures in removing secondary physical effects, even when masked-region fidelity remains high. These findings expose a substantial gap between visual plausibility and causal scene consistency, highlighting the need for evaluation frameworks that capture physical reasoning rather than local reconstruction alone.

The main contributions of this paper can be summarized as follows:
\begin{itemize}
    \item We introduce a causally grounded formulation of video object removal, reframing evaluation around object-environment consistency rather than masked-region reconstruction.
    \item We construct \emph{BeyondMasks}, a paired benchmark of 180 synthetic and real-world video sequences with aligned clean backgrounds, masks, and instruction prompts, spanning diverse causal interactions. It more than doubles the scale of prior paired video object removal benchmarks.
    \item We propose and validate \emph{CORE}, a structured VLM-based evaluation framework that jointly assesses object disappearance and the causal consistency of induced physical after-effects.
    \item We benchmark contemporary mask-based and instruction-driven video object removal models and reveal systematic failures in handling dynamic and interaction-driven physical effects.
\end{itemize}

\section{Related Work}
\label{sec:related_work}
\subsection{Video Object Removal Benchmarks}
Early evaluations of video object removal rely on segmentation datasets such as DAVIS~\cite{Perazzi_2016_CVPR} and YouTube-VOS~\cite{Xu_2018_ECCV}. Although these provide high-quality masks, they lack paired clean background videos in which both the object and its induced effects are absent. As a result, evaluation is limited to masked-region reconstruction, implicitly treating removal as local inpainting.
HQVI~\cite{cho2024rgvi} addresses this by constructing paired ground-truth videos via alpha compositing of foreground objects onto real backgrounds. While it provides temporally aligned pairs and occlusion-aware annotations, the compositing process assumes no secondary environmental effects, keeping evaluation focused on spatial reconstruction rather than object-environment disentanglement. ROSE-Bench~\cite{miao2025rose} introduces synthetic paired videos generated via 3D rendering and models side effects such as shadows, reflections, mirror interactions, light emission, and translucency\footnote{The authors also describe a real-world split generated via copy-and-paste compositing onto background videos, which is not publicly available at the time of writing.}. 
Despite advancing physically grounded evaluation, synthetic benchmarks remain limited in interaction diversity and environmental variability, primarily emphasizing isolated photometric effects. They do not systematically capture dynamic interactions such as deformation, fluid perturbations, particle traces, or scene rearrangement under realistic motion, nor do they support instruction-driven editing. 
Table~\ref{tab:dataset_comparison} summarizes the differences between prior benchmarks and BeyondMasks. In contrast, BeyondMasks provides temporally aligned paired videos across synthetic and real-world scenes, explicitly modeling diverse causal after-effects and dynamic interactions, enabling systematic evaluation beyond masked-region fidelity.

\begin{table}[!t]
\caption{\textbf{Comparison of video object removal benchmarks.} Unlike prior datasets, BeyondMasks provides temporally aligned paired ground truth while explicitly modeling diverse causal after-effects in both synthetic and real-world scenes.}\label{tab:dataset_comparison}
\tiny
\centering
\scriptsize
\setlength{\tabcolsep}{5pt}
\renewcommand{\arraystretch}{1.25}

\begin{tabular}{@{}l@{$\;\,$}c@{\;\,}c@{\;\,}c@{\;\,}c@{\;\,}c@{\;\,}c@{}}
\toprule
\textbf{Dataset} & \textbf{Size} & \textbf{Paired GT} & \textbf{After-effects} & \textbf{Causal effects} & \textbf{Avg. video length} & \textbf{Source} \\
\midrule
DAVIS & 90 & \No & \No & \No & $\sim$3 secs  & Real \\
YouTube-VOS & 508 & \No & \No & \No & $\sim$4.5 secs & Real \\
HQVI & 10 & \Yes & \No & \No & $\sim$3.5 secs & Real \\
ROSE-Bench & 60 & \Yes & \Yes & \No & $\sim$3 secs & Synthetic \\
Ours & 180 & \Yes & \Yes & \Yes & $\sim$5.5 secs & Both \\
\bottomrule
\end{tabular}
\end{table}

\subsection{Evaluation of Physical and Causal Consistency}
Video editing is typically evaluated using reference-based metrics such as PSNR, SSIM~~\cite{wang2004image}, and LPIPS \cite{zhang2018unreasonable}, which measure pixel similarity to ground truth. While effective for reconstruction quality, these metrics do not capture whether object-induced effects persist or whether dynamic physical interactions remain temporally consistent. Perceptual measures such as FVD \cite{unterthiner2019fvd} assess realism but also remain insensitive to intervention correctness. At the image level, task-specific metrics such as ReMOVE~\cite{Chandrasekar_2024_CVPR} and CFD~\cite{Yu_2025_ICCV} incorporate object-aware signals to penalize residual foreground traces and assess semantic consistency. However, they operate on single frames and do not address temporal coherence or dynamic object–environment interactions.
Recent benchmarks employ pretrained vision-language models (VLMs) as automated evaluators. VBench~\cite{vbench,vbench2}, VideoPhy~\cite{videophy,videophy2}, and LLM-based judge frameworks~\cite{llm_judge_survey} use structured prompting to assess semantic alignment and motion realism in generated videos. Yet these approaches evaluate realism without aligned references and are not designed for intervention-based editing. Causal object removal requires comparison against temporally aligned clean backgrounds to verify both object disappearance and elimination of induced physical effects. We address this through CORE, a structured VLM-based evaluation protocol that leverages paired references and explicitly reasons about object removal and residual after-effects. CORE enables scalable evaluation of causal consistency beyond masked-region fidelity.

\section{BeyondMasks: Benchmark Design and Construction}
\label{sec:dataset}

We introduce \emph{BeyondMasks}, a benchmark for evaluating video object removal under a causal intervention framework.  While the recent ROSE-Bench~\cite{miao2025rose} dataset incorporates selected physical side effects in synthetic settings, prior benchmarks largely assess removal through masked-region reconstruction or effect-specific realism. We instead formalize object removal as an intervention on a scene-generating process, aiming to recover the interventional scene state in which both the object and its induced environmental changes are absent, so that the resulting video corresponds to the scene as if the object had never been present.

Let $S_t$ denote the latent physical state of a scene at time $t$, and let $O$ denote the presence of a target object. We conceptualize each observed frame as arising from a rendering process $V_t = R(S_t, O)$, where a scene state together with the object configuration is mapped to pixel observations. Under an intervention in which the object is removed (i.e., $O = \varnothing$), the same scene state is rendered without the object, yielding the clean background frame $V_t^{bg} = R(S_t, \varnothing)$. This formulation highlights that object removal is not merely masked reconstruction, but a counterfactual rendering problem: recovering the scene as it would appear under the intervention $O = \varnothing$. Importantly, $O$ influences $V_t$ not only through direct appearance (object pixels), but also indirectly through environmental interactions that alter $S_t$, including shadows, reflections, illumination shifts, surface deformation, and dynamic perturbations. True object removal therefore requires approximating the interventional distribution $R(S_t, \varnothing)$ rather than merely hallucinating plausible background content within a masked region.

\begin{figure}[!t]
\centering
\includegraphics[width=1.0\linewidth]{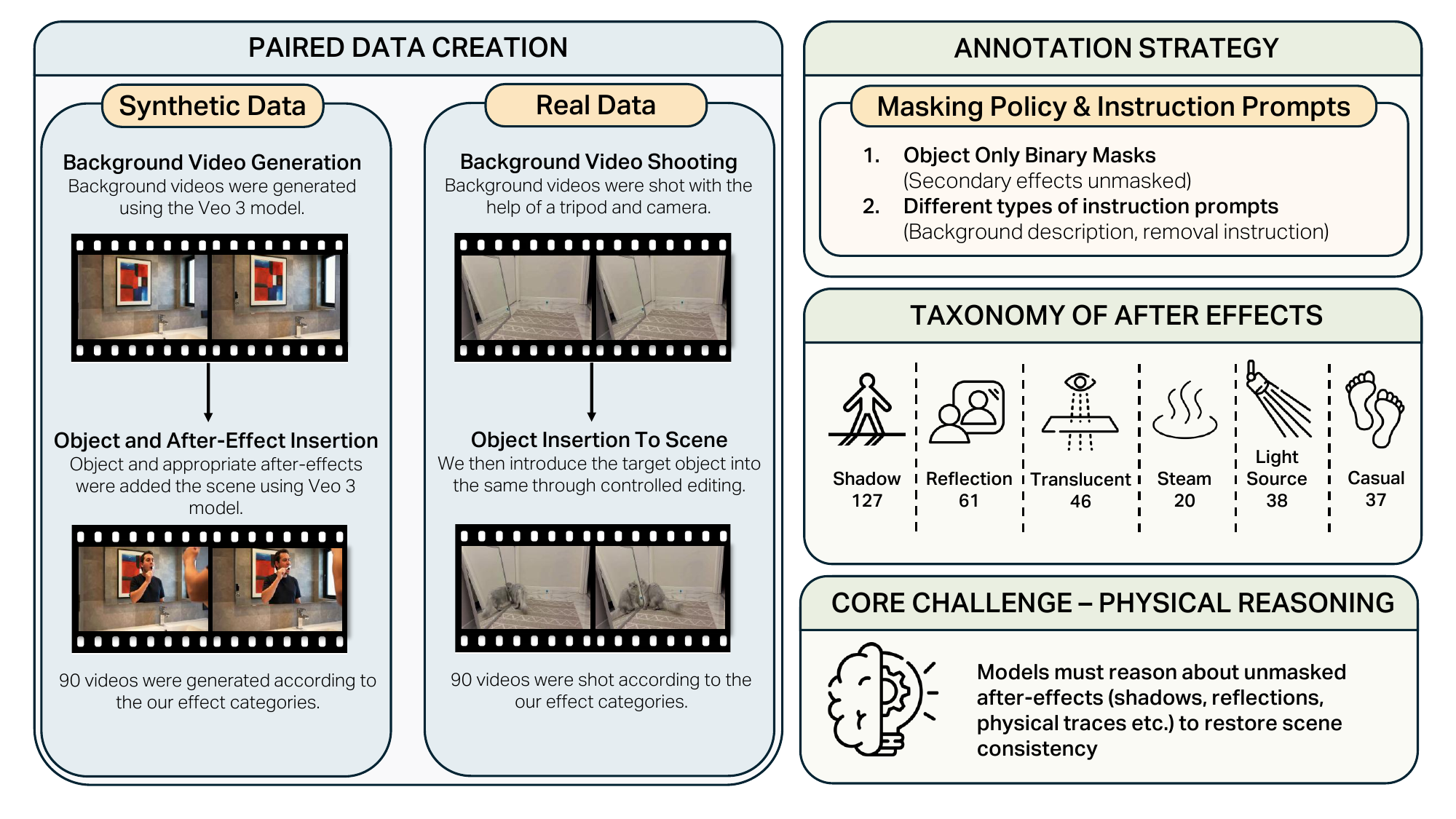}
\caption{\textbf{Overview of the BeyondMasks construction pipeline.} Top: Paired intervention-based generation, where an object-present video is derived from a clean background video, ensuring temporal alignment. Bottom: Representative categories of object-induced causal after-effects, including illumination, reflection, volumetric, and dynamic interaction effects.}
\label{fig:beyondmasks_overview}
\end{figure}

BeyondMasks implements this formulation through paired, temporally aligned intervention data, as illustrated in Fig.~\ref{fig:beyondmasks_overview}.  Each sample consists of:
(i) an object-present sequence $V^{obj}$ generated under $O \neq \varnothing$, and 
(ii) a clean reference sequence $V^{bg}$ generated under $O = \varnothing$.
Because the two sequences differ only by this controlled intervention, evaluation directly measures whether a removal model successfully recovers the interventional scene state. 

\subsection{Synthetic Paired Generation}
The synthetic subset contains 90 scenes generated using Veo3’s \cite{veo3} video editing capabilities. For each scene, we first generate a background-only video without the target object. This sequence serves as the clean reference. We then introduce the target object into the same scene through controlled editing, producing the object-present video. Since the latter is derived from the former, temporal alignment is preserved.
The insertion process naturally induces physically coherent interactions with the environment, including cast shadows, illumination changes, reflections, translucency effects, and scene-dependent dynamic traces. Because the background sequence remains unchanged apart from the object insertion, the paired design ensures that the only differences between videos correspond to the object and its causal footprint. This controlled construction enables both pixel-level comparison and higher-level semantic assessment.
All synthetic pairs undergo manual review to verify physical plausibility, temporal coherence, and consistency between object presence and environmental interaction.

\subsection{Real-World Paired Capture}
To complement synthetic data, BeyondMasks includes 90 real-world paired sequences captured with a tripod-stabilized setup. For each scene, we record a short video with the object present and immediately capture a second video after removing the object, keeping camera position or settings fixed. This back-to-back protocol preserves geometry, lighting, and scene composition, ensuring that differences between paired videos arise solely from object presence and its associated effects.
These real-world captures introduce natural texture complexity, subtle lighting variation, and realistic interaction patterns that are difficult to model synthetically, improving the realism and diversity of the benchmark.

\subsection{Mask Annotation Pipeline}
For each object-present sequence, we provide temporally consistent binary masks to support mask-based removal methods. Masks are generated through a semi-automatic pipeline that combines interactive segmentation using SAM2~\cite{sam2} with temporal propagation via CoTracker 3~\cite{cotracker3}. Annotators refine the initial-frame segmentation using positive and negative point prompts to obtain precise object boundaries, after which masks are propagated across frames and periodically corrected to maintain temporal coherence. Each sequence undergoes cross-review by a second annotator to ensure segmentation accuracy and consistency.
Importantly, masks include only the target object and explicitly exclude its after-effects. Shadows, reflections, illumination changes, and other induced phenomena are not masked. This design choice reflects our causal formulation: models must infer and remove secondary effects through physical reasoning rather than relying on explicit supervision.

\subsection{Taxonomy of Causal After-Effects}
A central component of BeyondMasks is its taxonomy of object-induced environmental interactions. Rather than categorizing scenes by object type, we group samples according to the underlying physical mechanisms through which objects causally influence their surroundings. This mechanism-driven taxonomy ensures that evaluation reflects distinct modes of object-environment entanglement.

\vspace{0.5em}\noindent\textbf{Shadow.}
Objects occlude or redirect illumination, producing cast shadows and local radiance changes. Successful removal requires restoring a shadow-consistent illumination field, not merely deleting object pixels.

\vspace{0.5em}\noindent\textbf{Reflection.}
Objects may appear indirectly through specular or diffuse reflections on mirrors, glass, or water. Removal must eliminate both direct and reflected evidence while preserving surface appearance and temporal consistency.

\vspace{0.5em}\noindent\textbf{Translucent effects.}
Semi-transparent objects (e.g., glass) partially reveal the background while modulating light transport. These cases require recovering background structure consistent with partial visibility and attenuation.

\vspace{0.5em}\noindent\textbf{Steam/scattering.}
Volumetric media such as steam or smoke introduce spatially diffuse, time-varying traces. Removal demands undoing these effects while maintaining coherent scene dynamics and background continuity.

\vspace{0.5em}\noindent\textbf{Causal physical effects.}
Some objects alter scene geometry or material state through motion or contact, yielding effects that are not visually attached to the object yet remain causally dependent on it. Examples include water ripples, dust trails, displaced objects, and surface deformation such as footprints.

\vspace{0.5em}\noindent\textbf{Light source effects.}
When the object emits light, its removal requires compensating for localized illumination changes and restoring consistent exposure and shading.

\vspace{0.5em}\noindent\textbf{No explicit after-effect.}
A small subset of sequences exhibits minimal secondary interaction, serving as controls for evaluating pure object removal with limited environmental entanglement.

\subsection{Instruction Prompts}
To support instruction-driven editing, each sample is associated with one or more validated text prompts specifying the removal target while emphasizing natural scene restoration. Prompts are constructed using structured templates to ensure object specificity and clarity, and are manually verified for consistency across both mask-based and text-conditioned paradigms.

\section{CORE: Causal Object Removal Evaluation}
\label{sec:vlm_eval}

\begin{figure}[!t]
\centering
\includegraphics[width=\linewidth]{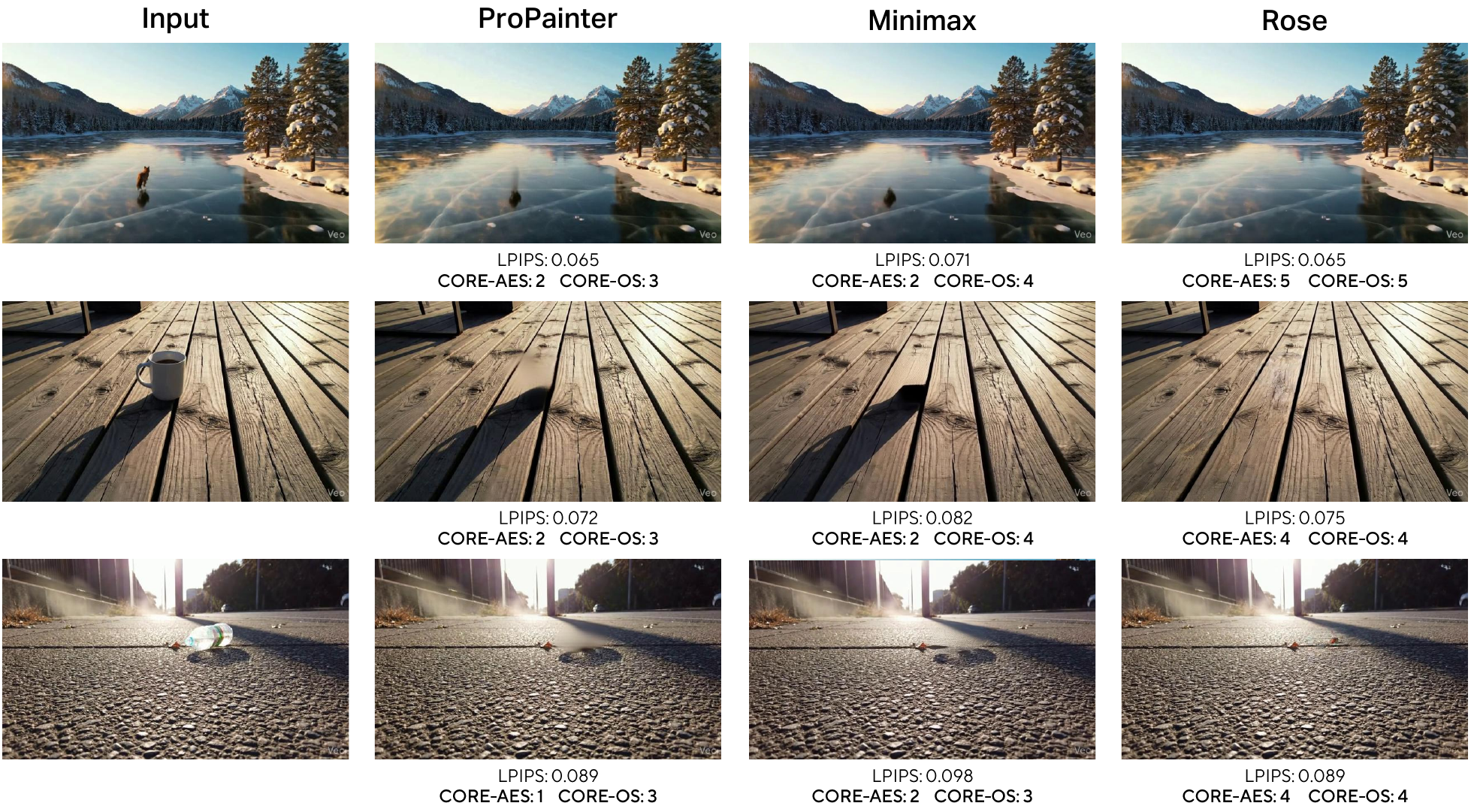}
\caption{\textbf{Pixel metrics fail to capture causal correctness.}
Although LPIPS scores are similar across methods, their ability to remove object-induced effects (e.g., shadows or reflections) differs substantially. CORE separates these failure modes by evaluating object removal (CORE-OS) and elimination of causal after-effects (CORE-AES). $\downarrow$ better for LPIPS; $\uparrow$ better for CORE scores.
}
\label{fig:core_motivation}
\end{figure}
\begin{figure}[!t]
\centering
\includegraphics[width=1.0\linewidth]{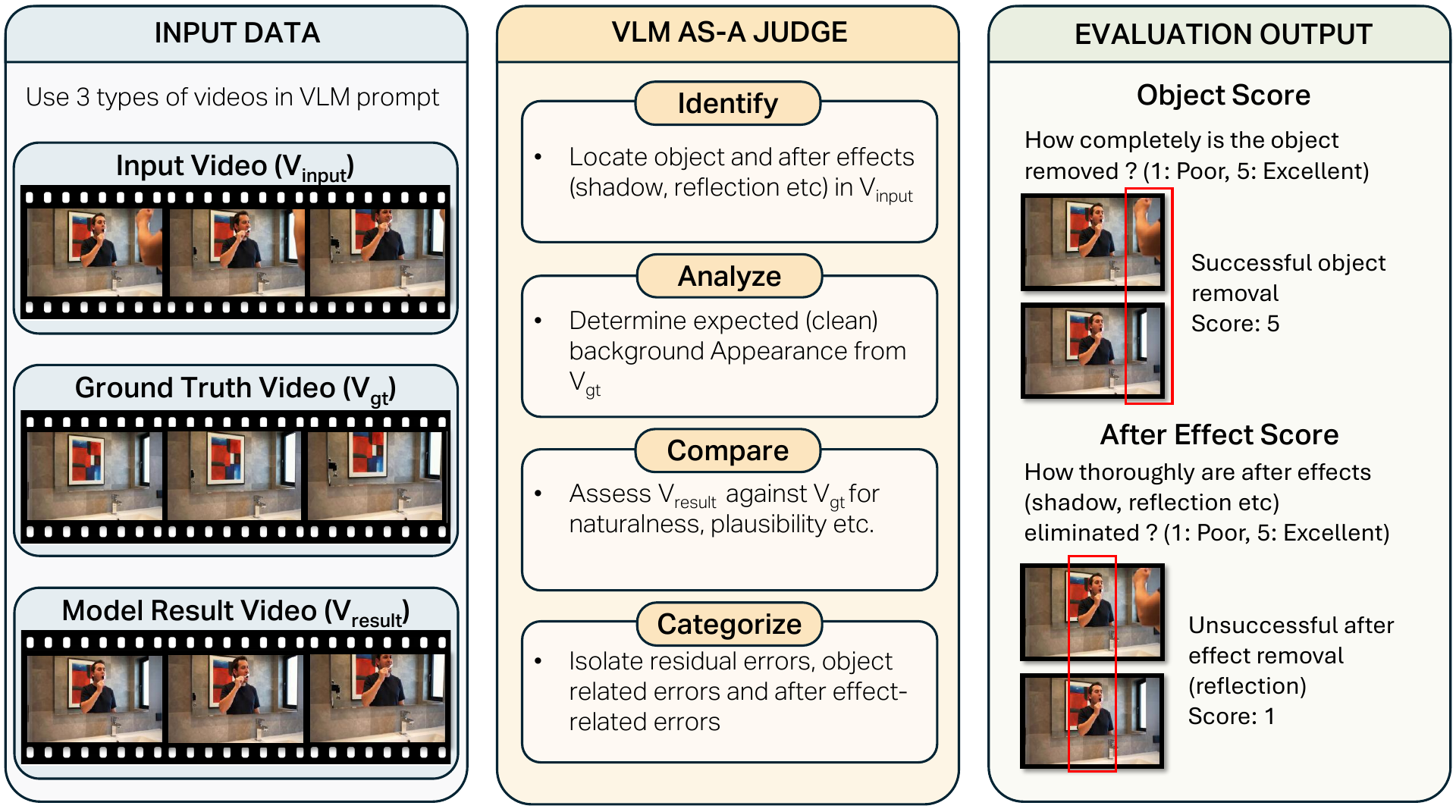}
\caption{\textbf{Overview of CORE (Causal Object Removal Evaluation)}. Given an edited video $\hat{V}^{bg}$ and its temporally aligned clean reference $V^{bg}$, CORE performs structured VLM-based comparison to assess interventional correctness. The evaluation jointly measures (i) object disappearance and (ii) consistency of object-induced physical after-effects, such as shadows, reflections, and dynamic perturbations. By conditioning semantic reasoning on aligned references, CORE moves beyond masked-region fidelity toward evaluation of causal and physical consistency.}
\label{fig:core_framework}
\end{figure}

While BeyondMasks enables paired pixel-level comparison, reconstruction metrics alone are insufficient for evaluating interventional correctness. Pixel-based measures such as PSNR, SSIM, and LPIPS penalize any deviation from the ground-truth background, even when the reconstructed scene is physically plausible and free of residual object traces. As illustrated in Fig.~\ref{fig:core_motivation}, methods with nearly identical LPIPS scores can differ substantially in whether object-induced physical effects are properly removed. In object removal, multiple visually coherent reconstructions may exist for unseen regions, making strict pixel identity an unreliable proxy for causal success.
Under our causal formulation, successful removal corresponds to recovering the interventional scene state $R(S_t, \varnothing)$, that is, the scene that would be observed if the object and its induced environmental changes were absent. Deviations from this target state arise at two distinct levels: (i) \emph{residual object presence}, where visible traces of the object remain, and (ii) \emph{residual causal effects}, where shadows, reflections, illumination shifts, or dynamic perturbations persist despite the object's removal.
To explicitly measure these dimensions, we introduce \emph{CORE (Causal Object Removal Evaluation)}, a structured VLM-based assessment framework illustrated in Fig.~\ref{fig:core_framework}. 

CORE evaluates removal using three temporally aligned inputs: the object-present video $V$, the aligned clean background $V^{bg}$, and the edited output $\hat{V}^{bg}$ produced by the evaluated method. 

Rather than measuring pixel correspondence, CORE assesses whether the edited video is semantically and physically consistent with the interventional reference in which the object and its induced environmental changes are absent. The evaluation is implemented via a pretrained vision-language model operating under a fixed, structured prompt that enforces comparative reasoning between $\hat{V}^{bg}$ and $V^{bg}$. The prompt explicitly guides the model to reason about two intervention-specific criteria: (i) whether the object has been fully removed, and (ii) whether its causal after-effects have been eliminated.
These two dimensions correspond to complementary failure modes in object removal. Residual object traces may remain even if the background appears plausible, and conversely, shadows or dynamic perturbations may persist despite apparent object disappearance. To disentangle these behaviors, CORE decomposes evaluation into two scores:
\begin{itemize}
    \item \textbf{ObjectScore:} completeness of object removal and plausibility of background restoration;
    \item \textbf{AfterEffectScore:} elimination of object-induced physical traces, including shadows, reflections, illumination changes, and dynamic perturbations.
\end{itemize}

We implement CORE using Gemini 3.1 \cite{deepmind2026gemini31pro}. Each evaluation call receives the three videos along with structured metadata specifying the target object and relevant after-effect categories. The VLM performs structured comparative reasoning: identifying object traces in the object-present video $V$, inferring expected appearance from the aligned clean background $V^{bg}$, comparing the edited output $\hat{V}^{bg}$  against $V^{bg}$ under a naturalness criterion, and categorizing residual errors as object-related or aftereffect-related.
The scoring rubric explicitly rewards semantically coherent background reconstruction, even when pixelwise differences from GT exist, while penalizing residual object evidence, persistent after-effects, or newly hallucinated foreground content. By conditioning semantic reasoning on aligned references, CORE transforms VLM-based scoring from observational realism assessment into structured intervention verification. The full prompt and rubric are provided in the supplementary material.

\vspace{0.5em}\noindent\textbf{Validation of CORE.}
To assess alignment with human judgment, we conduct a user study in which 8 annotators independently evaluate results from 5 methods across 20 videos using the same scoring rubric. The Pearson correlation between CORE scores and the mean human ratings is 0.615 for \textit{ObjectScore} and 0.733 for \textit{AfterEffectScore}. For reference, human inter-rater agreement yields Pearson correlations of 0.693 and 0.785 for the same dimensions, respectively. These results indicate that CORE closely aligns with human perception and approaches the upper bound defined by human agreement.

\section{Experimental Analysis}
\label{sec:experiments}

\subsection{Experimental Setup}
We benchmark nine recent video object removal and video editing methods spanning mask-based (M), text-guided (T), and hybrid (T+M) paradigms. These include classical mask-guided inpainting approaches such as ProPainter~\cite{zhou2023propainter} and DiffuEraser~\cite{li2025diffueraser}, layer-based decomposition models including Generative Omnimatte~\cite{lee2025generative} and OmnimatteZero~\cite{samuel2025omnimattezero}, and recent large-scale editing systems such as VACE~\cite{jiang2025vace}, Lucy Edit~\cite{team2025lucy}, CoCoCo~\cite{zi2025cococo}, MiniMax~\cite{zi2025minimax}, and ROSE~\cite{miao2025rose}. Together, these methods represent the dominant paradigms currently used for video object removal and instruction-driven video editing. 
All methods are evaluated using publicly released implementations and default configurations at the native resolution of BeyondMasks. For each method, we report standard reference-based metrics, PSNR, SSIM, LPIPS, computed against the paired GT backgrounds, as well as FVD for temporal quality and our proposed VLM-based CORE ObjectScore (CORE-OS) and AfterEffectScore (CORE-AES). 

\subsection{Quantitative Results}
\begin{table}[!t]
\caption{\textbf{Quantitative comparison on our benchmark.} We evaluate representative video object removal methods using standard reconstruction metrics (PSNR, SSIM, LPIPS), the video realism metric FVD, and our VLM-based CORE evaluation. 
\textbf{Type} indicates the supervision required by each method: \textbf{T} denotes text-instruction-based editing, \textbf{M} denotes mask-guided removal, and \textbf{T+M} indicates methods that support both inputs. \textbf{Bold} indicates the best score.}
\centering
\setlength{\tabcolsep}{4pt}
\renewcommand{\arraystretch}{1.25}

\resizebox{\textwidth}{!}{%
\begin{tabular}{@{}l@{$\;\,$}c@{$\;\,$}c@{$\;\,$}c@{$\;\,$}c@{$\;\,$}c@{$\;\,$}c@{$\;\,$}c@{}}
\toprule
\textbf{Method} & \textbf{Type} & \textbf{PSNR}$\uparrow$ & \textbf{SSIM}$\uparrow$ & \textbf{LPIPS}$\downarrow$ & \textbf{FVD}$\downarrow$ & \textbf{CORE-OS}$\uparrow$ & \textbf{CORE-AES}$\uparrow$\\
\midrule
Lucy Edit (arXiv 2025) \cite{team2025lucy} & T & 18.5884 & 0.7239 & 0.2758 & 480.08 & 1.443 & 1.551 \\
VACE (ICCV 2025) \cite{jiang2025vace} & T+M & 19.9932 & 0.7856 & 0.2077 & 418.79 & 1.799 & 1.698 \\
CoCoCo (AAAI 2025) \cite{zi2025cococo} & T+M & 20.8565 & 0.7287 & 0.2404 & 381.84 & 1.986 & 2.043 \\
Gen. Omnimatte (CVPR 2025) \cite{lee2025generative} & T+M & 23.8106 & 0.8175 & 0.1968 & 250.76 & 3.771 & 3.021 \\
ProPainter (ICCV 2023) \cite{zhou2023propainter} & M & 23.5979 & 0.8429 & 0.1635 & 225.28 & 3.063 & 2.368 \\
DiffuEraser (arXiv 2025) \cite{li2025diffueraser} & M & \textbf{25.1880} & \textbf{0.8769} & \textbf{0.1124} & \textbf{164.26} & 3.494 & 2.625 \\
MiniMax (NeurIPS 2025) \cite{zi2025minimax} & M & 23.1954 & 0.8218 & 0.1682 & 203.64 & 3.631 & 2.553 \\
ROSE (NeurIPS 2025) \cite{miao2025rose} & M & 23.8062 & 0.8173 & 0.1659 & 316.04 & 3.784 & 2.920 \\
OmnimatteZero (SIGGRAPH 2025) \cite{samuel2025omnimattezero} & M & 22.9306 & 0.8127 & 0.1638 & 227.24 & 3.190 & 2.499 \\
EffectErase (CVPR 2026) \cite{fu2026effecterasejointvideoobject} & M & 24.1206 & 0.8473 & 0.1621 & 242.16 & \textbf{4.015} & \textbf{3.162} \\
\bottomrule

\end{tabular}%
}
\label{tab:quant_results}
\end{table}

Table~\ref{tab:quant_results} summarizes the results. Across standard reconstruction metrics, several methods achieve comparable PSNR, SSIM, and LPIPS scores, indicating similar pixel-level fidelity with respect to the clean reference videos. However, these metrics do not reliably reflect removal correctness. Methods that achieve strong reconstruction scores frequently leave residual object evidence or fail to remove induced physical effects such as shadows or reflections. This discrepancy is particularly evident when comparing reconstruction metrics with CORE scores. For example, DiffuEraser attains the highest PSNR and SSIM, yet exhibits noticeably lower AfterEffectScore than several competing methods, indicating incomplete removal of object-induced effects. Conversely, ROSE achieves the highest CORE-OS despite only moderate reconstruction metrics, suggesting more effective elimination of object traces even when pixel similarity is comparable. These results highlight a fundamental limitation of conventional evaluation: pixel-based fidelity does not necessarily correspond to causal correctness. By explicitly separating object disappearance (CORE-OS) from elimination of physical after-effects (CORE-AES), CORE reveals failure modes that remain hidden under traditional reconstruction metrics. This analysis confirms that realistic video object removal requires reasoning about object-environment interactions rather than merely reconstructing masked regions.

To further analyze performance across different physical phenomena, Table~\ref{tab:effect_breakdown} reports results aggregated by after-effect category. The breakdown reveals substantial variation in difficulty across interaction types. While shadows and reflections are handled relatively well on average, volumetric effects such as steam and dynamic interactions (e.g., footprints or surface disturbances) exhibit significantly lower CORE-AES scores. These results indicate that secondary physical processes extending beyond the masked region remain challenging for current removal methods. Overall, the analysis confirms that realistic object removal requires modeling object-environment interactions rather than relying solely on local reconstruction.

\begin{table}[!t]
\caption{\textbf{Average performance across after-effect categories.} 
Results are averaged across the top-performing six models for each after-effect type. 
Higher values are better for PSNR, SSIM, CORE-OS, and CORE-AES, while lower values indicate better performance for LPIPS.}
\centering
\footnotesize
\setlength{\tabcolsep}{4pt}
\renewcommand{\arraystretch}{1.25}

\begin{tabular}{@{}l@{$\;\,$}c@{$\;\,$}c@{$\;\,$}c@{$\;\,$}c@{$\;\,$}c@{}}
\toprule
\textbf{After-effect} 
& \textbf{PSNR}$\uparrow$ 
& \textbf{SSIM}$\uparrow$ 
& \textbf{LPIPS}$\downarrow$ 
& \textbf{CORE-OS}$\uparrow$ 
& \textbf{CORE-AES}$\uparrow$ \\
\midrule

Shadow & 23.6870 & 0.8298 & 0.1621 & 3.498 & 2.520 \\
Reflection & 23.9842 & 0.8558 & 0.1580 & 3.332 & 2.446 \\
Light source & 22.4233 & 0.8201 & 0.1849 & 3.554 & 2.248 \\
Steam & 22.8375 & 0.8466 & 0.1624 & 3.236 & 2.032 \\
Translucent & 22.9326 & 0.8156 & 0.2162 & 3.256 & 2.701 \\
Causal & 22.7912 & 0.8163 & 0.1867 & 3.486 & 2.141 \\

\bottomrule
\end{tabular}
\label{tab:effect_breakdown}
\end{table}

\subsection{Qualitative Analysis}
\begin{figure}[!t]
\centering
\includegraphics[width=1.0\linewidth]{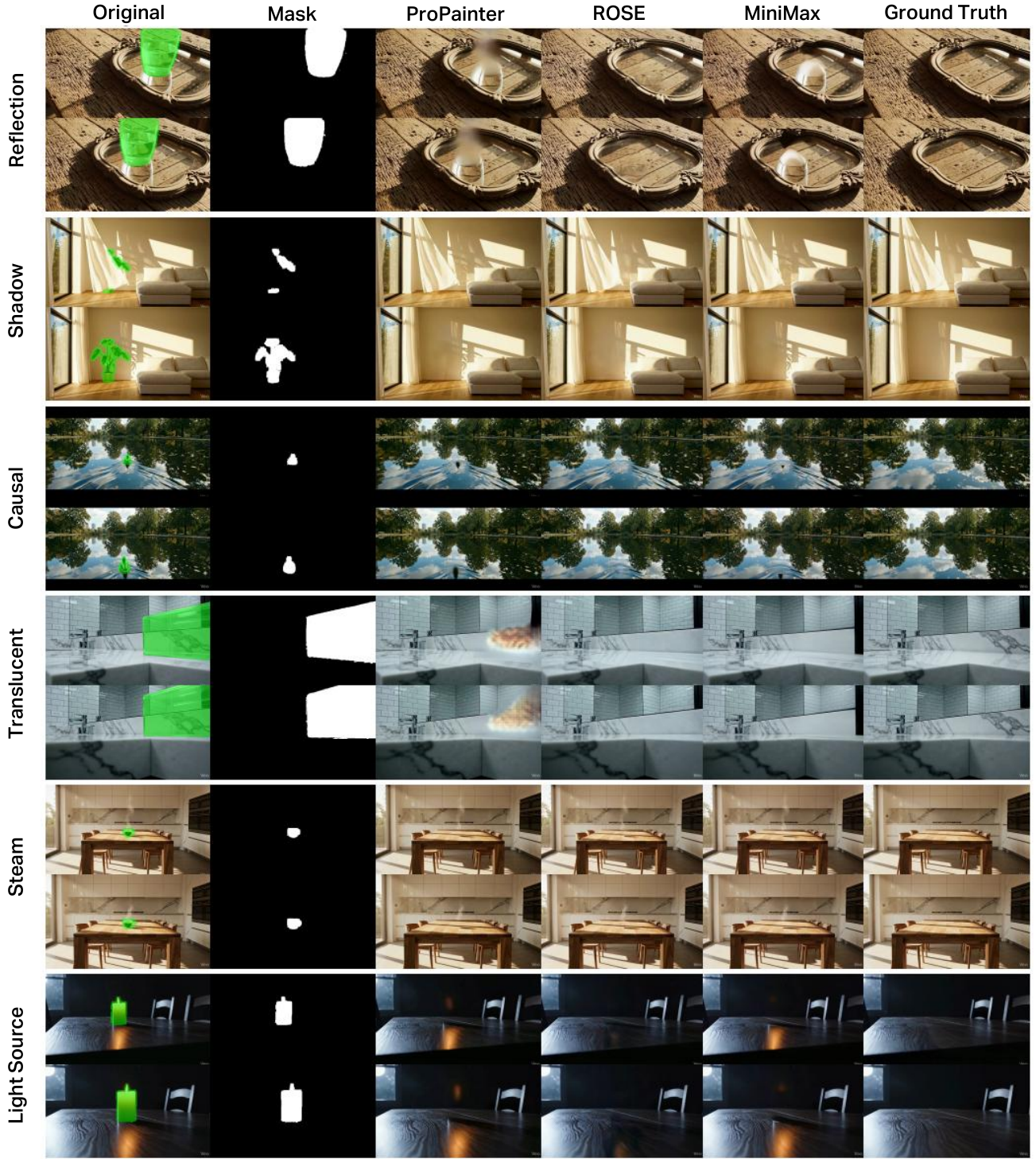}
\caption{\textbf{Qualitative comparison across six after-effect categories.}
Each category shows two temporal frames.}
\label{fig:qualitative_results}
\end{figure}

Fig.~\ref{fig:qualitative_results} presents representative examples comparing several state-of-the-art methods. While most approaches successfully remove the target objects, many fail to eliminate object-induced effects. Shadows, reflections, and illumination changes often persist even after the object disappears, resulting in physically inconsistent scenes. These residual traces are particularly evident in cases involving reflective surfaces, cast shadows, and lighting interactions. 

\vspace{0.5em}\noindent\textbf{After-Effect Removal is the Dominant Challenge.}
Across the examples in Fig.~\ref{fig:qualitative_results}, removing the object itself is often easier than eliminating the physical effects it induces in the environment. While most methods successfully erase the primary object, secondary phenomena such as shadows, reflections, and illumination changes frequently persist. These effects are spatially diffuse and may extend beyond the masked region, making them difficult to address with standard inpainting strategies. This observation highlights the importance of evaluating object removal as a causal intervention rather than a purely local reconstruction task.

\vspace{0.5em}\noindent\textbf{Mask-Based vs. Text-Guided Removal.}
Qualitative differences also emerge between mask-based and instruction-driven approaches. Mask-based methods typically achieve strong object removal due to explicit spatial supervision, but often leave residual environmental effects such as shadows or reflections. In contrast, text-guided approaches offer greater editing flexibility but show inconsistent removal of fine-grained physical traces and may introduce hallucinated content (see Supplementary for visual examples). These observations suggest that current systems, regardless of supervision modality, struggle to disentangle objects from the physical processes they induce in the surrounding environment.

\begin{figure}[!t]
    \centering
    \includegraphics[width=\textwidth]{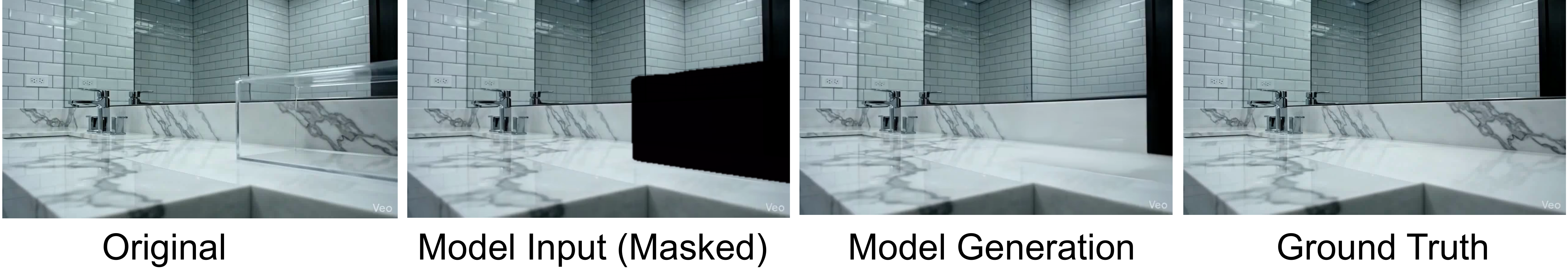}
    \caption{\textbf{Context loss in mask-based video object removal}. Many mask-based pipelines zero out the masked region prior to inference, eliminating potentially informative visual evidence (e.g., background partially visible through translucent structures). This removal of observable context forces the model to hallucinate missing content rather than recover available information, resulting in degraded restoration and incomplete causal consistency relative to the ground truth.}
    \label{fig:mask_problem}
\end{figure}

\vspace{0.5em}\noindent\textbf{Context Loss in Mask-Based Inpainting.}
A structural limitation of many mask-based pipelines is the removal or replacement of masked regions before inference. This operation discards potentially informative visual evidence, particularly for translucent or semi-transparent objects where background content remains partially observable. As illustrated in Fig.~\ref{fig:mask_problem}, eliminating these signals forces models to hallucinate missing content rather than recover available context. Similar observations have been reported in image-based removal~\cite{smarteraser}. In video, temporal redundancy provides additional opportunities to exploit such information across frames, suggesting that future removal systems should incorporate context-aware reasoning rather than relying solely on masked inpainting.

\section{Conclusion}
\label{sec:conclusion}
We introduced \emph{BeyondMasks}, a benchmark for evaluating video object removal from a causal intervention perspective. Unlike existing evaluation protocols that primarily assess masked-region reconstruction, BeyondMasks enables systematic evaluation of whether removal methods eliminate both the object and the physical effects it induces in the surrounding environment. By providing temporally aligned video pairs with clean background references and diverse interaction categories, the benchmark allows controlled analysis of object-environment disentanglement in realistic editing scenarios. Using this benchmark, we evaluated a diverse set of recent video object removal and video editing methods and identified a consistent gap between high reconstruction fidelity and true causal correctness. While many approaches successfully remove the visible object, they frequently leave behind residual shadows, reflections, illumination changes, or other secondary physical traces. To address limitations of existing evaluation protocols, we introduced \emph{CORE}, a VLM-based evaluation framework that separately measures object disappearance and the removal of object-induced after-effects. Our experiments show that conventional pixel-based metrics often fail to capture these failure modes, whereas CORE provides a more faithful assessment of causal consistency and aligns more closely with human perception.

\section*{Acknowledgements}
This work was partly supported by
the KUIS AI Center Research Awards to ES, and TÜB\.{I}TAK-2247-A Program Award (No.
123C550) to AE. We thank all the reviewers for their valuable comments.

\bibliographystyle{splncs04}
\bibliography{main}
\clearpage
\title{Supplementary Material for\\BeyondMasks: Evaluating Causal and Physical Consistency in Video Object Removal}
\titlerunning{BeyondMasks}

\author{
Yi\u{g}it Ekin\inst{1}\equalcontrib\orcidlink{0009-0006-4969-608X}
\and
Enes Sanli\inst{2,4}\equalcontrib\orcidlink{0009-0000-0361-6597}
\and
Aykut Erdem\inst{2,4}\orcidlink{0000-0002-6280-8422}
\and
Erkut Erdem\inst{3,4}\orcidlink{0000-0002-6744-8614}
\and
Aysegul Dundar\inst{1}\orcidlink{0000-0003-2014-6325}
}

\authorrunning{Y. Ekin et al.}

\institute{
Bilkent University, Department of Computer Engineering, Ankara, T\"urkiye
\and
Ko\c{c} University, Department of Computer Engineering, Istanbul, T\"urkiye
\and
Hacettepe University, Department of AI and Data Engineering, Ankara, T\"urkiye
\and
KUIS AI Center, Istanbul, T\"urkiye
\\
\email{yigit.ekin@bilkent.edu.tr} 
\\[0.4em]
\BeyondMasksLinks
}

\maketitle

\makeatletter
\renewcommand*{\theHsection}{supp.\arabic{section}}
\renewcommand*{\theHsubsection}{supp.\arabic{section}.\arabic{subsection}}
\makeatother

This supplementary document provides additional analyses, experimental results, and qualitative examples that complement the main paper. 
Sec.~\ref{sec:supp-dataset} describes the composition of the BeyondMasks dataset and the distribution of causal after-effect categories. 
Sec.~\ref{sec:video-removal} reviews representative video object removal approaches, while Sec.~\ref{sec:editing} discusses instruction-based video editing methods relevant to our evaluation setting. 
Sec.~\ref{sec:cfd-remove} evaluates existing image-based object removal metrics on the BeyondMasks benchmark and analyzes their limitations for video-based causal removal. 
Sec.~\ref{sec:split-results} reports additional quantitative results comparing performance on synthetic and real subsets of BeyondMasks. 

Sec.~\ref{sec:alignment} analyzes the alignment between CORE scores and human evaluations and examines the robustness of the VLM-based evaluation. 
Sec.~\ref{sec:prompt} provides the prompt template used in the CORE evaluation pipeline. 
Sec.~\ref{sec:phys-common} discusses the connection between BeyondMasks and physical commonsense evaluation. 
Sec.~\ref{sec:benchmark-diff} highlights the differences between BeyondMasks and existing benchmarks. 
Finally, Sec.~\ref{sec:qualitative} presents additional qualitative comparisons highlighting common failure cases.

\section{Dataset Composition}
\label{sec:supp-dataset}
BeyondMasks contains 180 paired video sequences, including 90 synthetic scenes and 90 real-world captures. Each sample consists of an object-present video, a temporally aligned clean background reference, per-frame object masks, and validated text prompts for instruction-driven removal. Although compact in size, the dataset is designed with controlled interventions and temporal alignment, allowing precise evaluation of whether both objects and their induced environmental effects are removed.

Synthetic sequences enable controlled modeling of diverse physical interactions, including shadows, reflections, translucent effects, and dynamic scene perturbations. Real-world captures complement these controlled scenes by introducing natural textures, lighting variations, and complex interactions that are difficult to reproduce synthetically. Together, this combination provides both controlled evaluation conditions and realistic scene complexity.

Table~\ref{tab:effect_distribution} summarizes the distribution of after-effect categories in BeyondMasks. Categories are not mutually exclusive: a single video may contain multiple interaction types, and therefore category counts do not sum to the total number of videos. The distribution reflects the natural frequency of object-induced visual effects in real scenes rather than an artificially balanced design. For example, shadows occur in many everyday lighting conditions, while phenomena such as steam, scattering, or strong translucent interactions appear less frequently. Preserving this natural distribution makes the benchmark more representative of realistic object removal scenarios.

\begin{table}
\centering
\caption{\textbf{Distribution of after-effect categories in BeyondMasks.} Categories are not mutually exclusive; a single video may exhibit multiple interaction types.}
\label{tab:effect_distribution}

\begin{minipage}[t]{0.45\linewidth}
\centering
\vspace{0pt}
\footnotesize
\begin{tabular}{l@{}c}
\toprule
After-Effect Category & Count \\
\midrule
Shadow & 127 \\
Reflection & 61 \\
Translucent effects & 46 \\
Causal physical effects & 37 \\
Light source effects & 38 \\
Steam / scattering & 20 \\
No explicit after-effect & 6 \\
\bottomrule
\end{tabular}
\end{minipage}
\hfill
\begin{minipage}[t]{0.45\linewidth}
\centering
\vspace{0pt}
\includegraphics[width=\linewidth]{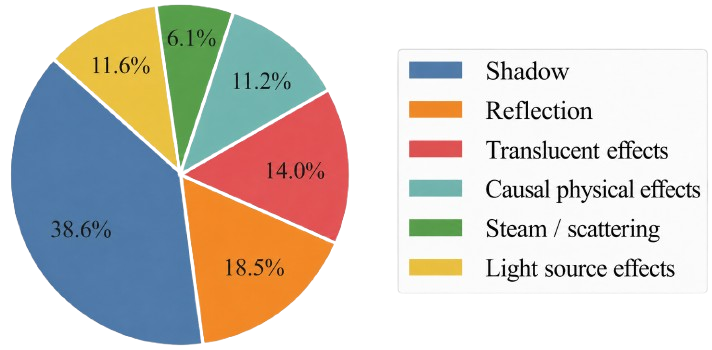}
\end{minipage}

\end{table}
\subsection{Validation of Background Preservation in Synthetic Pairs}
\label{sec:bg-preservation}

A key requirement of the synthetic subset is that each object-present video and its corresponding clean reference represent the same underlying scene under a controlled intervention. In other words, the paired videos should differ only in the presence of the target object and in the visual consequences causally induced by that object. Establishing this property is important because unintended changes in distant background regions would make it difficult to attribute differences between the input and reference videos to the object removal intervention itself.

Standard reconstruction metrics such as PSNR and SSIM are not well suited for validating this property when computed outside the target object mask. In our setting, the object mask deliberately covers only the visible object, while many relevant effects---including shadows, reflections, local illumination changes, steam, dust, water disturbances, and other interaction traces---remain outside the mask. These regions are expected to differ between the object-present input and the clean reference because they constitute part of the object's causal footprint.

\begin{figure}[!h]
\centering
\resizebox{\linewidth}{!}{  
\begin{tabular}{@{}c@{$\,$}c@{$\,$}c@{$\,\,$}c@{$\,$}c@{$\,$}c@{}}
\includegraphics[width=0.16\linewidth]{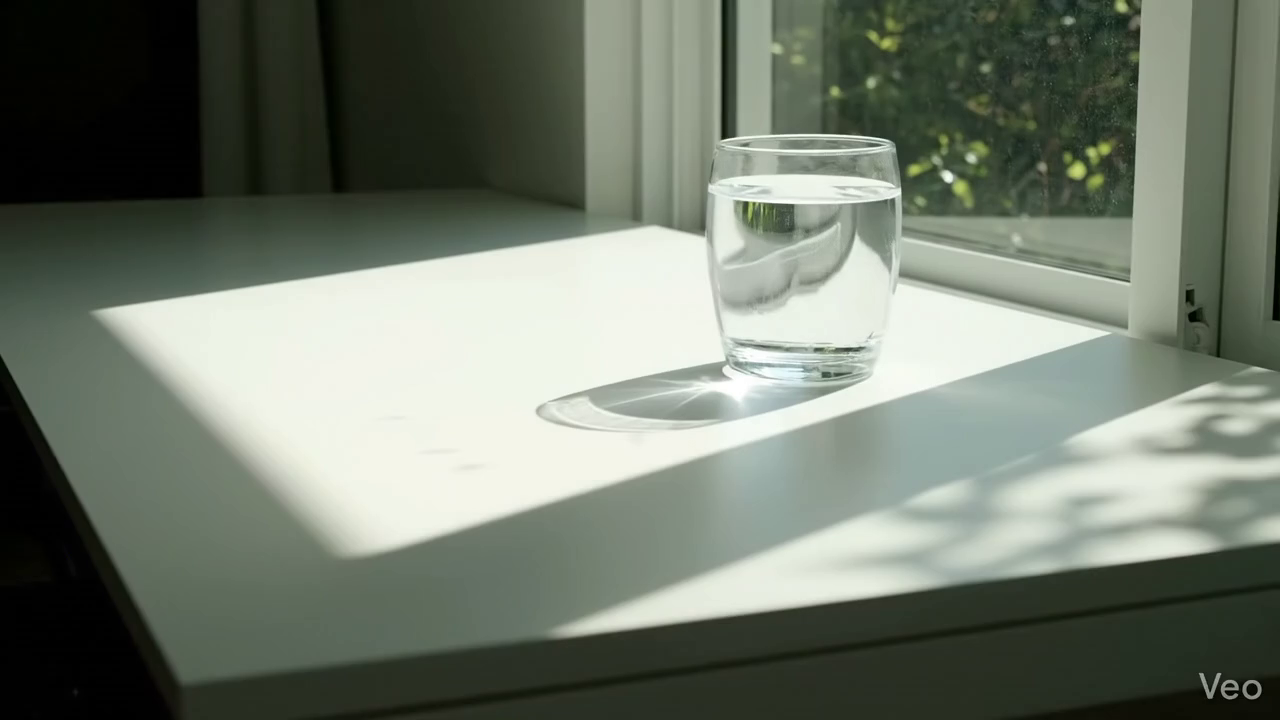} &
\includegraphics[width=0.16\linewidth]{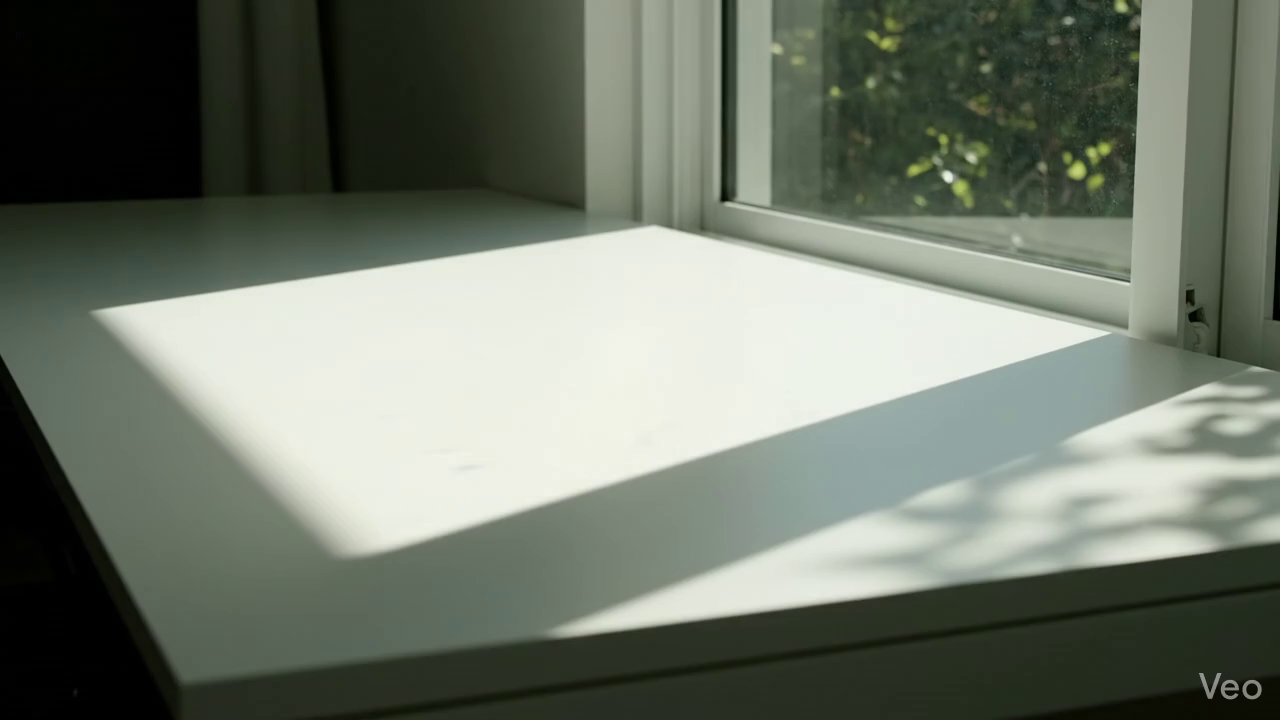} &
\includegraphics[width=0.16\linewidth]{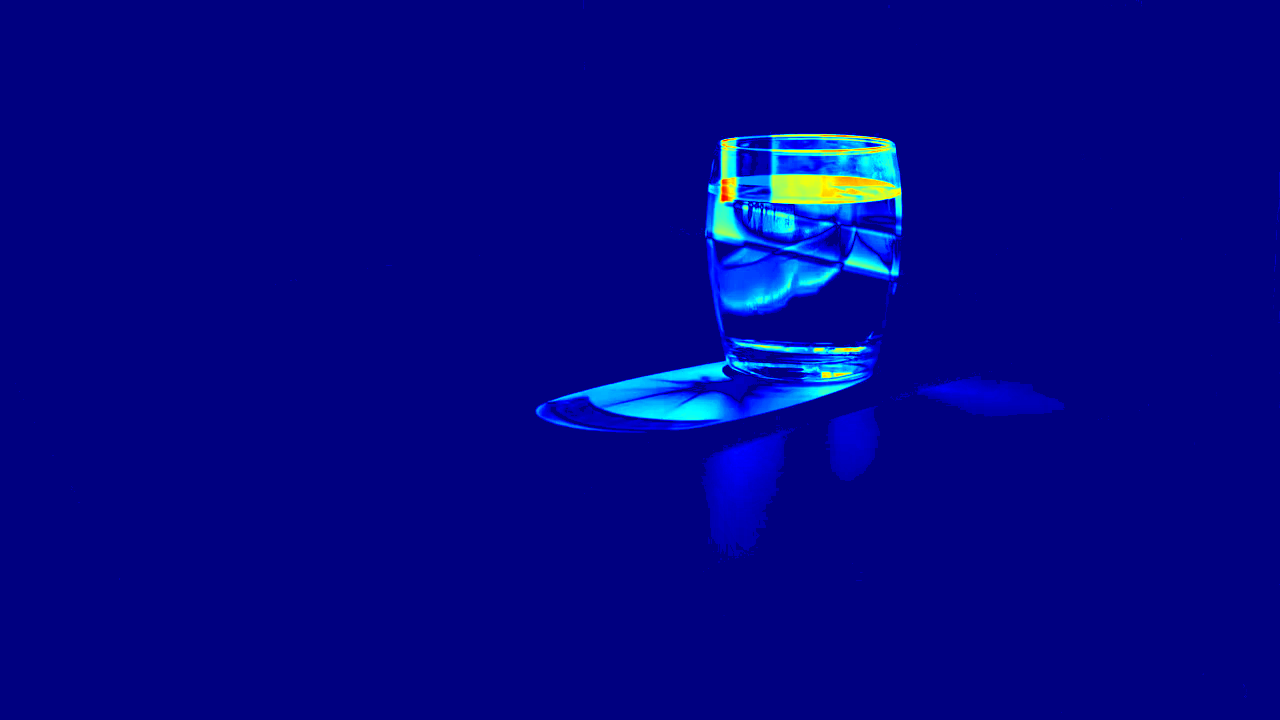} &

\includegraphics[width=0.16\linewidth]{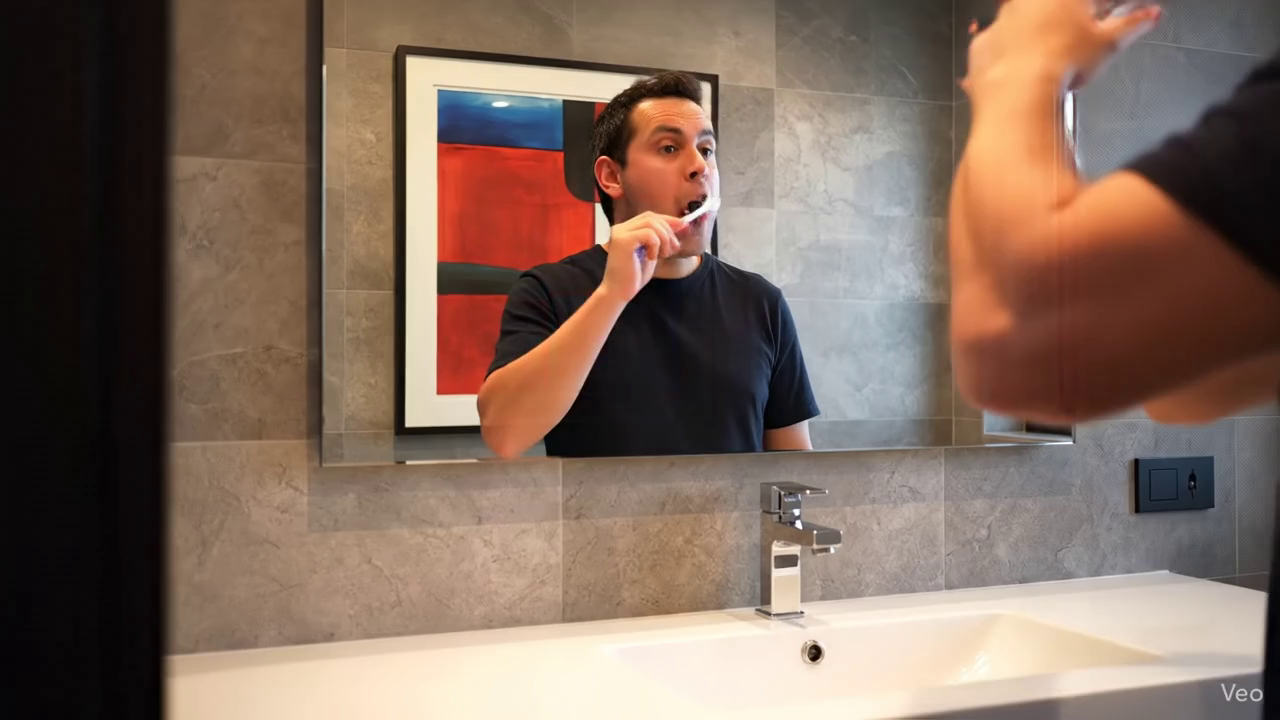} &
\includegraphics[width=0.16\linewidth]{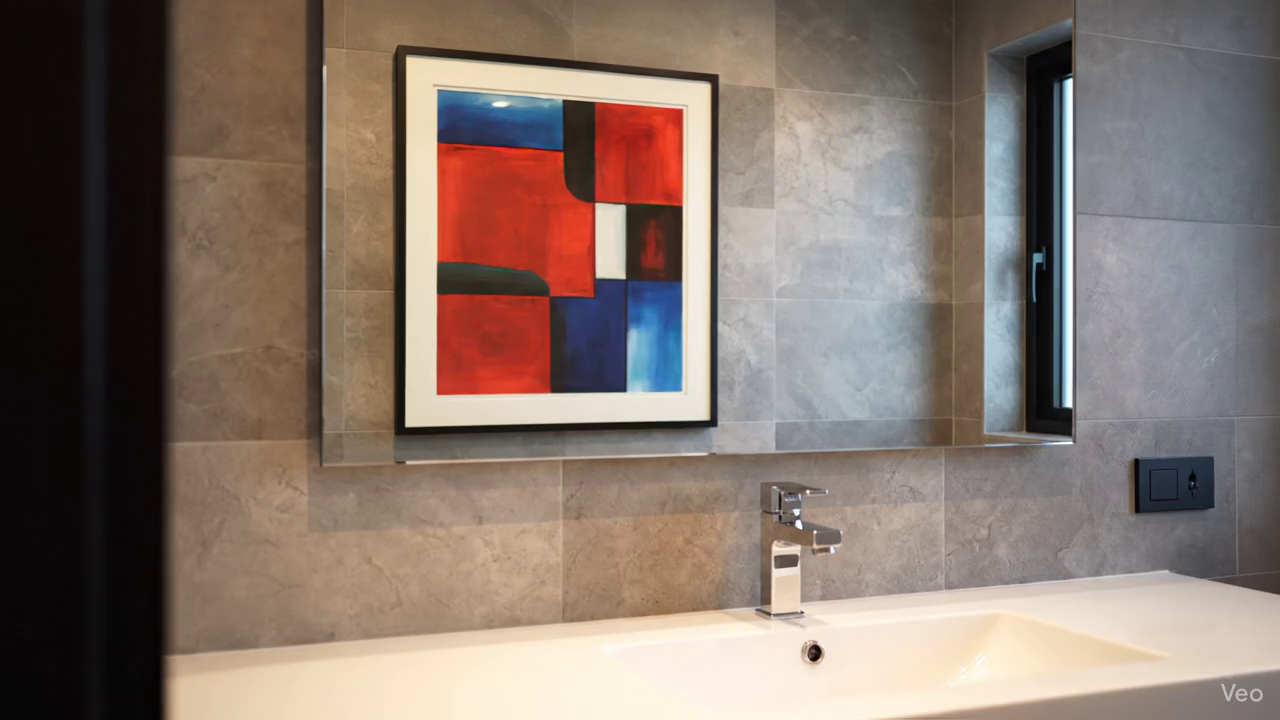} &
\includegraphics[width=0.16\linewidth]{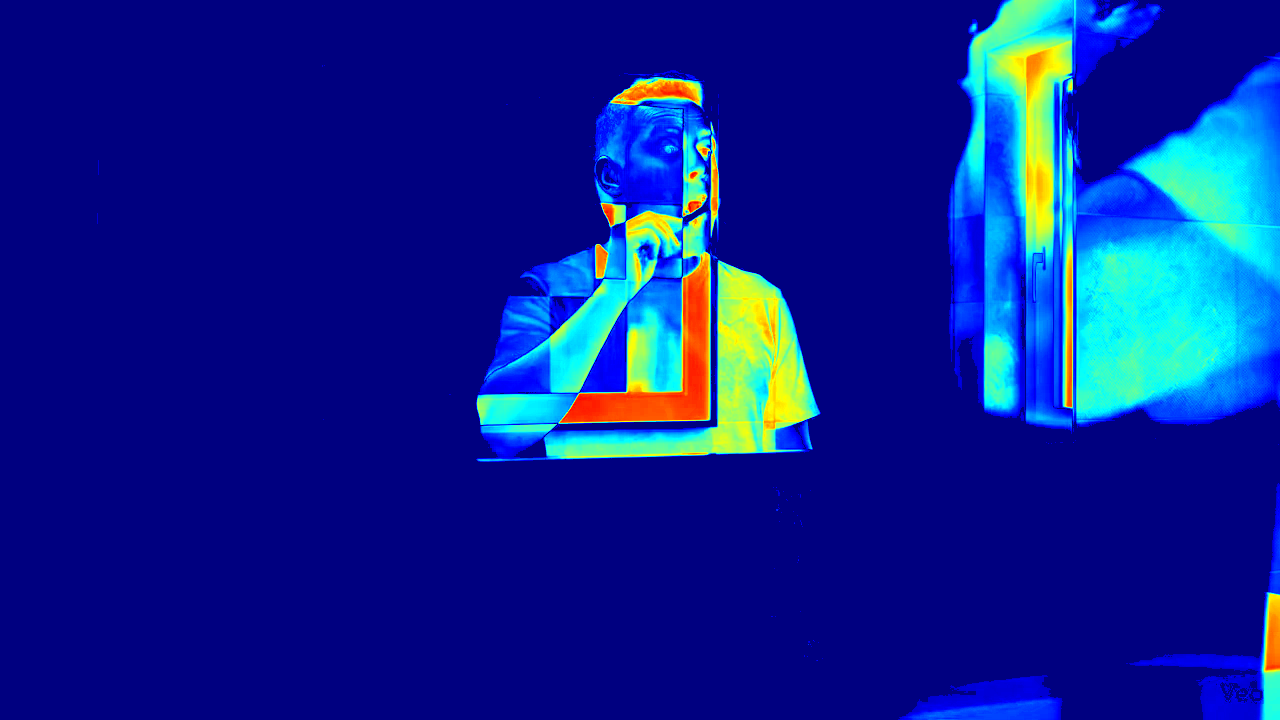} \vspace{-0.15cm}\\
{\scriptsize Input} & {\scriptsize Ground Truth} & {\scriptsize Difference} & {\scriptsize Input} & {\scriptsize Ground Truth} & {\scriptsize Difference}\vspace{-0.35cm}
\end{tabular}}
\caption{\textbf{Localized differences in synthetic pairs.}\vspace{-0.1em}}
\label{fig:diff}
\end{figure}

Instead, we validate the synthetic pairs through systematic inspection of absolute per-pixel difference maps between the object-present input and the corresponding clean reference. As illustrated in Fig.~\ref{fig:diff}, meaningful differences are concentrated around the inserted object and its associated physical footprint. For example, the difference maps may capture a cast shadow, a reflection, a local lighting variation, or a dynamic disturbance caused by the object. In contrast, static and spatially distant background regions should remain nearly unchanged.

\section{Video Object Removal Methods}
\label{sec:video-removal} 
Video object removal is commonly formulated as a spatiotemporal inpainting problem guided by object masks. Early approaches extend image inpainting techniques to videos using optical flow, patch-based synthesis, and temporal propagation to maintain cross-frame consistency ~\cite{zhou2023propainter}. More recent deep learning methods incorporate 3D convolutional architectures, attention mechanisms, and learned temporal priors to improve visual realism and temporal coherence \cite{zi2025cococo}. Representative systems such as ProPainter~\cite{zhou2023propainter} and DiffuEraser~\cite{li2025diffueraser} combine motion-aware propagation or diffusion-based synthesis to reconstruct plausible background content within masked regions.

A second line of work leverages diffusion-based video foundation models for editing tasks. These models enable masked video editing through latent-space manipulation and transformer-based architectures \cite{miao2025rose,zi2025minimax }. Large-scale pretraining improves perceptual realism and motion consistency across frames. However, these systems remain fundamentally mask-conditioned inpainting models: they are optimized to synthesize visually plausible background content rather than explicitly reason about or eliminate object-induced physical effects such as shadows, reflections, or illumination changes.

Another family of approaches relies on layered video decomposition. Methods such as Omnimatte~\cite{Lu_2021_Omnimatte} and its generative extension OmnimatteZero~\cite{samuel2025omnimattezero} model scenes as compositions of foreground layers and background content. By explicitly separating dynamic objects and their associated effects (e.g., shadows or reflections), these models support flexible video editing and compositing. Nevertheless, they are primarily designed for layer manipulation and scene editing, and they are not evaluated under standardized settings that test the complete removal of object-induced environmental effects.

Recent image inpainting research has begun to address object removal beyond simple mask completion by explicitly modeling the visual effects induced by objects. Several methods aim to jointly remove objects together with associated artifacts such as shadows, reflections, or illumination changes through effect-aware attention mechanisms or paired supervision~\cite{objectclear,omnieraser}. Other approaches formulate object removal as a counterfactual reconstruction problem or leverage geometry-aware scene representations to eliminate objects and their causal visual effects in a physically consistent manner~\cite{objectdrop, georemover}. While these advances improve semantic completeness in static images, they remain limited to single-frame settings and do not address the temporal persistence of object–environment interactions in videos.

Overall, existing approaches largely treat object removal as a reconstruction problem, optimizing masked-region fidelity and perceptual plausibility. As a result, evaluation protocols typically focus on pixel-level similarity or local inpainting quality rather than verifying whether object-induced physical effects have been fully eliminated. BeyondMasks addresses this gap by enabling systematic evaluation of object-environment disentanglement and the removal of induced physical effects in temporally consistent video sequences.

\section{Instruction-Based Video Editing and Removal Methods}
\label{sec:editing}
Instruction-based visual editing has recently emerged as a powerful paradigm for controllable content manipulation using natural language prompts. Early work such as \textsc{InstructPix2Pix}~\cite{brooks2023instructpix2pix} demonstrated that diffusion models can be trained to follow textual instructions for image editing tasks, including object modification, insertion, and removal. These models enable flexible editing without requiring explicit spatial masks, instead relying on language guidance to specify the desired transformation.

Building on these ideas, subsequent work has extended instruction-conditioned editing from images to videos. Methods such as \textsc{InstructVid2Vid}~\cite{qin2024instructvid2vid} adapt diffusion-based editing models to the temporal domain by introducing mechanisms that propagate edits consistently across frames. Similarly, \textsc{InstructVideo} \cite{yuan2024instructvideo} improves instruction following by aligning video diffusion models with human feedback, enabling more accurate interpretation of natural language editing commands. These approaches illustrate the growing capability of video foundation models to perform flexible, language-driven video manipulation.

Within this paradigm, object removal can be performed using natural language instructions (e.g., ``remove the cup from the table'') rather than explicit masks. Recent video editing systems therefore increasingly support instruction-driven removal alongside traditional mask-conditioned editing. For example, \textsc{VACE}~\cite{jiang2025vace} unifies multiple editing modes within a single architecture, including masked editing, reference-based generation, and instruction-guided video manipulation. Such frameworks highlight the trend toward integrating language interfaces into video editing pipelines, allowing users to specify complex edits using natural language prompts.

Despite these advances, evaluating instruction-based object removal remains challenging. Existing studies typically rely on qualitative inspection or general-purpose video generation metrics such as Fréchet Video Distance (FVD)~\cite{fvd} or perceptual similarity measures such as LPIPS~\cite{lpips}. While these metrics capture perceptual realism and distributional similarity, they do not explicitly verify whether the target object and its associated physical effects (e.g., shadows, reflections, or illumination changes) have been fully removed.

BeyondMasks addresses this limitation by enabling structured evaluation of instruction-driven removal alongside mask-conditioned approaches. The dataset provides temporally aligned pairs of object-present videos and corresponding clean background references, allowing edited outputs to be compared directly with the ground-truth interventional scene state. This setup enables systematic analysis of whether instruction-based editing systems remove both the target object and the physical effects it induces in the surrounding environment.

\section{Image Based Object Removal Metrics}
\label{sec:cfd-remove}
In addition to the evaluation metrics reported in the main manuscript, we analyze model performance using two representative image-based object removal metrics: ReMOVE~\cite{Chandrasekar_2024_CVPR} and CFD Score~\cite{Yu_2025_ICCV}. Both metrics were originally designed to evaluate object removal quality in static images. We apply them frame-wise to our video benchmark to investigate whether image-level evaluation metrics reflect removal quality in the presence of object-induced physical effects.

To compute these metrics on videos, we sample frames from each sequence with a stride of 15 to balance temporal coverage and computational cost. The metric is computed independently for each sampled frame and averaged across frames to obtain a video-level score. Finally, scores are aggregated across videos belonging to the same causal after-effect category to produce category-level results.

Tables~\ref{tab:remove} and~\ref{tab:cfd} report the performance of all evaluated methods across the different after-effect categories in BeyondMasks using ReMOVE and CFD, respectively. These metrics were originally proposed for object-aware image evaluation, and thus they are not designed to capture temporally consistent causal effects in videos. As a result, their rankings do not necessarily reflect whether object-induced environmental interactions—such as shadows, reflections, or illumination changes—have been correctly removed.

\begin{table*}[!h]
\caption{\textbf{Evaluation using the ReMOVE metric on BeyondMasks across causal after-effect categories.} Higher values indicate better performance. \textbf{Type} denotes the supervision modality: \textbf{T} indicates text-instruction-based editing, \textbf{M} indicates mask-guided removal, and \textbf{T+M} denotes methods supporting both inputs. Bold numbers indicate the best score in each column. \textbf{Bold} numbers indicate the best score in each column.}
\centering
\setlength{\tabcolsep}{4pt}
\renewcommand{\arraystretch}{1.2}

\resizebox{\textwidth}{!}{%
\begin{tabular}{@{}l@{$\;\,$}c@{$\;\,$}c@{$\;\,$}c@{$\;\,$}c@{$\;\,$}c@{$\;\,$}c@{$\;\,$}c}
\toprule
\textbf{Method} & \textbf{Type} & \textbf{Shadow} & \textbf{Reflection} & \textbf{Light} & \textbf{Steam} & \textbf{Translucent} & \textbf{Causal} \\
\midrule
Lucy Edit (arXiv) \cite{team2025lucy} & T & 0.659 & 0.647 & 0.586 & 0.636 & 0.675 & 0.669 \\
VACE (ICCV 2025) \cite{jiang2025vace} & T+M & 0.658 & 0.645 & 0.597 & 0.630 & 0.679 & 0.666 \\
CoCoCo (AAAI 2025) \cite{zi2025cococo} & T+M & 0.662 & 0.651 & 0.593 & \textbf{0.648} & 0.685 & 0.661 \\
Gen. Omnimatte (CVPR 2025) \cite{lee2025generative} & T+M & 0.676 & 0.663 & \textbf{0.622} & 0.638 & \textbf{0.705} & 0.684 \\
ProPainter (ICCV 2023) \cite{zhou2023propainter} & M & 0.668 & 0.669 & 0.614 & 0.636 & 0.701 & 0.665 \\
DiffuEraser (arXiv) \cite{li2025diffueraser} & M & 0.664 & 0.667 & 0.615 & 0.647 & 0.696 & 0.679 \\
MiniMax (NeurIPS 2025) \cite{zi2025minimax} & M & 0.667 & 0.655 & 0.603 & 0.636 & 0.690 & 0.686 \\
ROSE (NeurIPS 2025) \cite{miao2025rose} & M & \textbf{0.678} & \textbf{0.670} & 0.595 & 0.636 & 0.704 & 0.690 \\
OmnimatteZero (SIGGRAPH 2025) \cite{samuel2025omnimattezero} & M & 0.669 & 0.611 & 0.601 & 0.617 & 0.699 & \textbf{0.698} \\
\bottomrule
\end{tabular}%
}
\label{tab:remove}
\end{table*}

\begin{table*}[!h]
\caption{\textbf{Evaluation using the CFD metric on BeyondMasks across causal after-effect categories.} Higher values indicate better performance. \textbf{Type} denotes the supervision modality: \textbf{T} indicates text-instruction-based editing, \textbf{M} indicates mask-guided removal, and \textbf{T+M} denotes methods supporting both inputs. Bold numbers indicate the best score in each column. \textbf{Bold} numbers indicate the best score in each column.}
\centering
\setlength{\tabcolsep}{4pt}
\renewcommand{\arraystretch}{1.2}

\resizebox{\textwidth}{!}{%
\begin{tabular}{@{}l@{$\;\,$}c@{$\;\,$}c@{$\;\,$}c@{$\;\,$}c@{$\;\,$}c@{$\;\,$}c@{$\;\,$}c}
\toprule
& & \multicolumn{6}{c@{$\;\,$}}{\textbf{CFD Score}$\uparrow$} \\
\cmidrule(lr){3-8}
\textbf{Method} & \textbf{Type} & \textbf{Shadow} & \textbf{Reflection} & \textbf{Light} & \textbf{Steam} & \textbf{Translucent} & \textbf{Causal} \\
\midrule
Lucy Edit (arXiv) \cite{team2025lucy} & T & \textbf{0.677} & \textbf{0.652} & \textbf{0.667} & \textbf{0.763} & \textbf{0.805} & \textbf{0.668}\\
VACE (ICCV 2025) \cite{jiang2025vace} & T+M & 0.558 & 0.585 & 0.520 & 0.631 & 0.670 & 0.594 \\
CoCoCo (AAAI 2025) \cite{zi2025cococo} & T+M & 0.326 & 0.358 & 0.368 & 0.317 & 0.450 & 0.292 \\
Gen. Omnimatte (CVPR 2025) \cite{lee2025generative} & T+M & 0.285 & 0.350 & 0.359 & 0.376 & 0.404 & 0.260\\
ProPainter (ICCV 2023) \cite{zhou2023propainter} & M & 0.253 & 0.248 & 0.341 & 0.275 & 0.301 & 0.249 \\
DiffuEraser (arXiv) \cite{li2025diffueraser} & M & 0.291 & 0.314 & 0.396 & 0.391 & 0.332 & 0.260 \\
MiniMax (NeurIPS 2025) \cite{zi2025minimax} & M & 0.285 & 0.318 & 0.351 & 0.404 & 0.301 & 0.257 \\
ROSE (NeurIPS 2025) \cite{miao2025rose} & M & 0.287 & 0.304 & 0.343 & 0.410 & 0.336 & 0.306 \\
OmnimatteZero (SIGGRAPH 2025) \cite{samuel2025omnimattezero} & M & 0.359 & 0.360 & 0.361 & 0.422 & 0.421 & 0.354 \\
\bottomrule
\end{tabular}%
}
\label{tab:cfd}
\end{table*}

Consistent with the observations in the main paper, the results indicate that image-based metrics often assign similar scores to methods that exhibit substantially different levels of causal removal quality. Because ReMOVE and CFD operate as no-reference metrics, they primarily evaluate semantic plausibility and object absence within individual frames rather than verifying, against reference videos, whether object-induced environmental effects have been removed. As a result, methods that leave residual reflections, shadows, or illumination changes may still obtain competitive scores if the edited frame appears visually coherent.

More specifically, ReMOVE tends to behave like a coarse scene-level plausibility metric, producing very similar scores across methods even when their actual removal quality differs substantially. CFD focuses more on visual compatibility within the edited frame and can therefore assign high scores to outputs that remain harmonious with the surrounding scene despite incomplete removal. Consequently, methods that leave clear residual object traces or after-effects may still appear competitive under CFD if the frame remains visually coherent. These observations highlight an important limitation of existing image-based metrics when applied to video object removal tasks involving complex object–environment interactions.

\section{Quantitative Results on Synthetic and Real Subsets}
\label{sec:split-results}
Table~\ref{tab:supp-quant-results} presents split-wise quantitative results on the synthetic and real subsets of BeyondMasks. The synthetic subset contains 90 videos generated under controlled intervention settings, while the real subset consists of 90 tripod-captured scenes. Reporting results separately allows us to assess whether trends observed in controlled environments transfer to real-world scenarios. Overall, performance patterns remain largely consistent across the two splits: methods that perform strongly on synthetic data generally remain competitive on real videos as well.

\begin{table}[!h]
\caption{\textbf{Split-wise quantitative evaluation on BeyondMasks.} Performance of representative video object removal and image object removal methods on the synthetic and real subsets of the dataset. Each entry reports \textit{Synthetic / Real} results for reconstruction metrics (PSNR, SSIM, LPIPS) and the proposed CORE evaluation scores (CORE-OS and CORE-AES), which are measured on a discrete 1–5 scale. Image-based methods (OmniPaint and Object Clear) are applied frame-wise by processing one frame every 15 frames. \textbf{Bold} indicates the best value for each metric across all methods.}
\centering
\setlength{\tabcolsep}{4pt}
\renewcommand{\arraystretch}{1.25}

\resizebox{\textwidth}{!}{%
\begin{tabular}{@{}l@{$\;\,$}c@{$\;\,$}c@{$\;\,$}c@{$\;\,$}c@{$\;\,$}c@{}}
\toprule
\textbf{Method} & \textbf{PSNR}$\uparrow$ & \textbf{SSIM}$\uparrow$ & \textbf{LPIPS}$\downarrow$ & \textbf{CORE-OS}$\uparrow$ & \textbf{CORE-AES}$\uparrow$\\

\midrule
Lucy Edit (arXiv) \cite{team2025lucy}
& 18.8356 / 18.3282
& 0.7529 / 0.6942
& 0.2228 / 0.3285
& 1.069 / 1.813
& 1.276 / 1.824 \\

VACE (ICCV 2025) \cite{jiang2025vace}
& 18.9362 / 18.6913
& 0.7547 / 0.7236
& 0.2209 / 0.3010
& 1.415 / 2.176
& 1.404 / 1.989 \\

CoCoCo (AAAI 2025) \cite{zi2025cococo}
& 21.6499 / 20.1249
& 0.7625 / 0.6950
& 0.1962 / 0.2809
& 1.670 / 2.297
& 1.875 / 2.209 \\

Gen. Omnimatte (CVPR 2025) \cite{lee2025generative}
& 25.0218 / 22.6120
& 0.8561 / 0.7794
& 0.1369 / 0.2562
& 3.865 / \textbf{3.670}
& \textbf{2.966} / \textbf{3.055} \\

ProPainter (ICCV 2023) \cite{zhou2023propainter}
& 24.6675 / 22.6555
& 0.8816 / 0.8060
& 0.1202 / 0.2044
& 3.326 / 2.800
& 2.371 / 2.361 \\

DiffuEraser (arXiv) \cite{li2025diffueraser}
& \textbf{25.3617} / \textbf{25.1622}
& \textbf{0.8932} / \textbf{0.8621}
& \textbf{0.0942} / \textbf{0.1290}
& 3.640 / 3.344
& 2.461 / 2.794 \\

MiniMax (NeurIPS 2025) \cite{zi2025minimax}
& 24.1725 / 22.2273
& 0.8597 / 0.7844
& 0.1164 / 0.2195
& 3.775 / 3.506
& 2.382 / 2.692 \\

ROSE (NeurIPS 2025) \cite{miao2025rose}
& 24.9255 / 22.6790
& 0.8558 / 0.7786
& 0.1128 / 0.2191
& \textbf{4.068} / 3.505
& 2.932 / 2.912 \\

OmnimatteZero (SIGGRAPH 2025) \cite{samuel2025omnimattezero}
& 22.9243 / 23.0160
& 0.8225 / 0.8040
& 0.1420 / 0.1834
& 3.438 / 2.945
& 2.360 / 2.638 \\

\midrule

OmniPaint (ICCV 2025) \cite{yu2025omnipaintmasteringobjectorientedediting}
& 23.6996 / 24.2534
& 0.8529 / 0.8436
& 0.1121 / 0.1512
& 3.750 / 3.125
& 3.125 / 2.182 \\

Object Clear (CVPR 2026) \cite{zhao2025objectclearcompleteobjectremoval}
& 25.5278 / 24.5341
& 0.8720 / 0.8532
& 0.0936 / 0.1326
& 4.325 / 3.725
& 3.022 / 3.015 \\

\bottomrule
\end{tabular}%
}
\label{tab:supp-quant-results}
\end{table}

Considering reconstruction-based metrics, DiffuEraser achieves the strongest overall performance, obtaining the highest PSNR and SSIM values together with the lowest LPIPS on both subsets. However, high reconstruction fidelity does not necessarily imply correct object removal. To evaluate removal correctness, we report the proposed CORE scores: \textit{ObjectScore} (CORE-OS) and \textit{AfterEffectScore} (CORE-AES). These scores are measured on a discrete 1–5 scale and reflect VLM-based judgments of object disappearance and the removal of object-induced effects, respectively.

The CORE scores reveal different performance trends than reconstruction metrics. For example, ROSE achieves the highest CORE-OS on the synthetic subset (4.068), indicating strong object disappearance under controlled conditions. However, its score decreases on the real subset (3.505), suggesting that removal becomes more challenging in scenes with richer visual interactions. Generative Omnimatte and DiffuEraser obtain the highest CORE-AES values among video-based methods, although the absolute scores remain moderate on the 1–5 scale, highlighting the difficulty of eliminating subtle physical after-effects such as shadows, reflections, and secondary interactions.

To provide additional reference points, we also evaluate two recent image-based object removal methods, OmniPaint~\cite{yu2025omnipaintmasteringobjectorientedediting} and ObjectClear~\cite{zhao2025objectclearcompleteobjectremoval}. Since these models operate on single images rather than videos, we apply them frame-wise by processing one frame every 15 frames. Interestingly, these methods achieve competitive reconstruction metrics and relatively strong CORE scores on individual frames. For instance, ObjectClear obtains the highest CORE-OS values across both subsets, indicating that modern image editing models can effectively remove objects and their associated effects at the frame level.

Nevertheless, image-based methods lack explicit temporal modeling and therefore cannot enforce consistency across frames. While frame-wise editing can locally remove objects and their visual artifacts, dedicated video models remain necessary to maintain temporal coherence and handle persistent object–environment interactions over time.

Overall, the split-wise results indicate that the synthetic subset captures performance trends that largely transfer to real scenes, while the real subset confirms that causally consistent object removal remains challenging in visually complex environments.

\section{Why Current Removal Methods Fail on Causal After-Effects}
\label{sec:failure-causes}

The qualitative examples reveal that the main limitation of current video object removal methods is not simply imperfect reconstruction inside the object mask. Rather, most failures arise because existing methods implicitly formulate removal as a spatially localized inpainting problem. Given a mask indicating the visible object, the model is typically optimized to synthesize plausible content within or near that masked region. However, an object's visual influence is often distributed beyond its visible boundary. Shadows, reflections, illumination changes, dust trails, water disturbances, steam, and other interaction traces may extend far outside the mask and can persist over time even after the object itself is no longer visible.

The difficulty varies across after-effect categories. Cast shadows are relatively more tractable because they are often spatially adjacent to the object and can be approximated as local photometric changes. Nevertheless, soft or temporally varying shadows may still extend beyond the model's effective editing region. Reflections are more challenging because their appearance depends on scene geometry, material properties, viewpoint, and the motion of both the object and the camera. Dynamic effects such as steam, dust, smoke, water ripples, or displaced particles are particularly difficult because they can spread over large regions and evolve according to temporal physical processes rather than remaining attached to a fixed object boundary.

Translucent objects introduce an additional ambiguity. Unlike opaque objects, they do not fully block the background; instead, each observed pixel may contain a mixture of foreground appearance and partially visible background content. A binary mask treats these mixed pixels as fully unavailable, discarding useful observations of the background behind the object. As a result, a mask-conditioned method must hallucinate details that are only partially occluded in the original frame. Recovering these details reliably requires exploiting temporal redundancy, since background content may be more clearly visible in earlier or later frames as the object or camera moves.

These observations suggest three design principles for causally consistent video object removal. First, models require scene-level representations rather than relying solely on mask-local features, because the relevant evidence may be spatially distant from the target object. Second, removal systems should preserve and aggregate partially observed background information across time, especially for translucent objects and thin structures. Third, models should explicitly account for object-conditioned lighting and temporally persistent effects instead of treating every out-of-mask change as fixed background. Together, these requirements indicate that causally consistent removal is fundamentally a counterfactual scene reasoning problem rather than a conventional masked reconstruction task.
\section{Human Alignment and Cross-VLM Robustness of the CORE Evaluation Metric}
\label{sec:alignment}
To assess how well automatic metrics reflect human perception of video object removal quality, we measure the correlation between metric outputs and human ratings. We report Spearman rank correlations between metric scores and the average human annotations across videos for both evaluation dimensions: \textit{ObjectScore} and \textit{AfterEffectScore}. The results are summarized in Table~\ref{tab:metric_corr}.

\begin{table}[!h]
\caption{\textbf{Correlation with human evaluation.} Spearman correlation between automatic metrics and human ratings for ObjectScore and AfterEffectScore. Pixel-based metrics (PSNR, SSIM, LPIPS) are compared with VLM-based CORE evaluations using two judge models (Gemini and GPT). The \textit{Inter Rater} row reports agreement between human annotators, serving as a reference level for achievable alignment.}
\centering
\footnotesize
\setlength{\tabcolsep}{4pt}
\renewcommand{\arraystretch}{1.25}

\begin{tabular}{@{}l@{$\;\,$}c@{$\;\,$}c@{}}
\toprule
\textbf{Metric} 
& \textbf{ObjectScore}
& \textbf{AfterEffectScore}  \\
\midrule
PSNR          & 0.15 & 0.28 \\
SSIM          & -0.14 & -0.10 \\
LPIPS         & -0.02 & 0.10 \\
CORE$_{\text{Gemini}}$        &  0.62 & 0.73  \\
CORE$_{\text{GPT}}$           &  0.60 & 0.76  \\
\midrule
Inter Rater           &  0.69 & 0.79  \\
\bottomrule
\end{tabular}
\label{tab:metric_corr}
\end{table}

Standard pixel-based reconstruction metrics exhibit weak correlation with human judgments. PSNR shows only limited agreement with human ratings, while SSIM and LPIPS display near-zero or negative correlations for both scoring dimensions. These results suggest that reconstruction fidelity alone does not reliably capture the aspects of removal quality that humans consider important. In particular, pixel similarity does not necessarily reflect whether the target object has been fully removed or whether object-induced environmental effects, such as shadows, reflections, or illumination changes, have been eliminated. Consequently, methods with similar reconstruction scores may still differ substantially in their causal correctness.

In contrast, the proposed CORE evaluation demonstrates substantially stronger agreement with human assessments. CORE$_{\text{Gemini}}$ achieves correlations of 0.62 and 0.73 for ObjectScore and AfterEffectScore, respectively, while CORE$_{\text{GPT}}$ obtains 0.60 and 0.76. The consistency of these results across two independent VLM backbones indicates that the effectiveness of CORE is not tied to a specific judge model. Instead, the improvement appears to stem from the structured evaluation protocol itself, which explicitly compares edited videos against aligned reference backgrounds and separates object removal from the elimination of causal after-effects.

We further examine the stability of the VLM-based evaluation by repeating the Gemini-based scoring multiple times and measuring agreement across runs using the intraclass correlation coefficient (ICC). The resulting ICC values are 0.808 for ObjectScore and 0.882 for AfterEffectScore, indicating strong run-to-run consistency. These results suggest that CORE produces stable and reliable judgments, particularly for evaluating the removal of residual causal traces.

\section{CORE Evaluation Prompt}
\label{sec:prompt}
CORE evaluates object removal using a structured prompt that guides a vision-language model (VLM) to compare the edited video against the aligned reference background. The same prompt template is used for all samples to ensure consistent evaluation across methods. Only two fields are instantiated for each example: \texttt{\{fg\_object\}}, which specifies the target object to be removed, and \texttt{\{focus\_effect\}}, which indicates the relevant after-effect categories (e.g., shadow, reflection, illumination change).

For each evaluation instance, the VLM receives three videos: (1) the \emph{INPUT} video containing the target object and its associated effects, (2) the \emph{GT} video representing the ideal background with both the object and its effects removed, and (3) the \emph{RESULT} video produced by the evaluated method. The prompt instructs the VLM to perform structured reasoning before assigning scores. Specifically, the model first identifies the object and its effects in the INPUT video, then analyzes the background appearance in the GT reference, and finally compares the RESULT video against the GT to determine whether the object and its induced effects have been successfully removed.

CORE decomposes evaluation into two complementary scores: \emph{ObjectScore}, which measures the completeness of object removal and plausibility of the reconstructed background, and \emph{AftereffectScore}, which measures whether object-induced visual effects such as shadows, reflections, or lighting changes have been eliminated. Both scores follow a 1-5 scale, where higher values indicate better removal quality. Requiring structured reasoning encourages the VLM to explicitly analyze object disappearance and residual environmental effects before producing the final scores.

Listing~\ref{lst:vlm_prompt} shows the prompt template used in our evaluation pipeline.

\begin{lstlisting}[
  style=promptstyle,
  caption={Prompt template used for CORE-based evaluation.},
  label={lst:vlm_prompt}
]
You are an expert computer vision judge evaluating a video object removal task using THREE separate videos.
The three videos provided represent:
INPUT Video: The original reference video where the target object and its aftereffects ARE PRESENT.
GT Video (Ground Truth): The IDEAL result video where the object and aftereffects are perfectly removed, showing the true background across all frames.
RESULT Video: Our method's output video. Your goal is to evaluate how well the RESULT video removes the target object and reconstructs a plausible, natural background.

Context for this sample:
Target object (present in INPUT; absent in GT): {fg_object}
Aftereffect types to check: {focus_effect}

EVALUATION PROCESS (THINKING STEPS)
Before scoring, you MUST output a step-by-step reasoning process using the following structure:
Step 1: Target Identification: Watch the INPUT video and identify the exact appearance, movement, and location of the Target Object ({fg_object}) and its Aftereffects ({focus_effect}).
Step 2: Expected Background Pattern (GT Analysis): Watch the GT video. Understand the general context, texture, and lighting of the background where the object used to be.
Step 3: Output vs GT Comparison (Naturalness Check): Compare the RESULT video against the GT. Evaluate the inpainted area. Does the filled background look natural and fit the surrounding scene harmoniously? Did the model successfully create a clean background, or did it hallucinate a new foreground object?
Step 4: Error Categorization: Determine which identified differences are related to the main Target Object removal, and which are related to the Aftereffect removal. Find the specific flaws based on the rubrics below.

SCORING RUBRICS (1-5 Scale)
HUMAN-LIKE JUDGING GUIDELINES:
- Humans care most about SEMANTIC removal. If the identifiable features of the target object are gone, humans consider the removal mostly successful (Score 4), EVEN IF a blurry smudge or a dark moving blob is left behind.
- A score of 3 should be reserved for partial temporal failures or when a small piece of the object remains.
- Punish heavily (Scores 1-2) ONLY if the object is fully visible for a long time, if the model hallucinates a completely new recognizable object, or if the artifact left behind is grotesquely out of place.

ObjectScore Rubric (How well the main target object was removed):
5 Points: Perfect or Near Perfect. The object is completely removed. The inpainted background seamlessly matches the GT. Minor, almost unnoticeable imperfections are acceptable.
4 Points: Good (Forgiving of Smudges). The core semantic features of the target object are completely gone. The inpainted area might contain noticeable smudges, blurriness, or dark moving blobs, but it does not severely break the overall geometry of the scene. The human eye forgives these as long as the object itself is unrecognizable.
3 Points: Fair (Partial/Distracting). The removal is flawed but not a total failure. Use this if: 1) A small but recognizable part of the object remains. 2) Temporal failure: The object is removed, but suddenly reappears for a few frames. 3) The object is gone, but the artifact left behind is highly distracting and physically illogical for the scene.
2 Points: Poor. Huge remnants of the target object are still clearly visible and moving. OR the model hallucinated a completely new, identifiable foreground object that completely ruins the background.
1 Point: Fail. The object is practically untouched and fully visible for the majority of the video, or the generated artifacts cause extreme, video-breaking corruption.

AftereffectScore Rubric (How well object-caused effects like shadows/reflections were removed):
5 Points: Perfect. The aftereffect is completely removed, matching the GT seamlessly.
4 Points: Good. Aftereffect is successfully removed and the area looks natural, even if lighting/texture slightly differs from the exact GT.
3 Points: Fair. Aftereffect area is noticeably blurry, smudged, or improperly lit compared to the rest of the scene.
2 Points: Poor. Severe visual artifacts in the aftereffect region.
1 Point: Fail. The aftereffect is still clearly visible in the RESULT video, completely ignored by the removal process.

OUTPUT FORMAT
You MUST format your response exactly like this:
Reasoning:
Step 1: [Your reasoning]
Step 2: [Your reasoning]
Step 3: [Your reasoning]
Step 4: [Your reasoning]
AftereffectScore: X, ObjectScore: Y
\end{lstlisting}

\section{Relation to Physical Commonsense Evaluation}
\label{sec:phys-common}
BeyondMasks is closely related to recent efforts aimed at evaluating physical commonsense understanding in video models. Benchmarks such as VideoPhy~\cite{videophy}, VideoPhy-2~\cite{videophy2}, and related evaluation frameworks have shown that videos that appear visually convincing can still violate basic physical expectations. These studies highlight a recurring gap between perceptual realism and genuine physical understanding in video generation systems.

Our findings in video object removal reveal a similar phenomenon from a different perspective. Many methods produce visually plausible outputs when evaluated with standard reconstruction or perceptual metrics, yet still fail to remove object-induced after-effects such as shadows, reflections, illumination changes, or motion traces. In these cases, the edited video may appear superficially realistic, but the model has not correctly accounted for how the removed object interacts with the surrounding scene.

From this perspective, BeyondMasks can be viewed as evaluating a form of \emph{counterfactual physical reasoning}. Rather than asking whether a generated video depicts a physically plausible event, our benchmark asks whether a model can correctly infer how a scene would appear if a particular object had never been present. Removing an object is therefore not merely a local appearance editing problem: the object may influence nearby pixels indirectly through lighting, geometry, material interactions, and temporal dynamics. Correctly editing the scene requires reasoning about these dependencies and eliminating their visual consequences.

This setting complements existing physical commonsense benchmarks. Prior benchmarks primarily evaluate whether models can \emph{generate} videos that obey intuitive physical behavior. BeyondMasks instead evaluates whether models can \emph{edit} an existing scene in a way that remains physically and causally consistent after an intervention. By providing paired object-present videos and clean reference backgrounds, the benchmark allows direct comparison between the edited result and the expected counterfactual scene state.

Together, these perspectives suggest that studying physical reasoning in video models should not be limited to forward generation tasks alone. Counterfactual editing tasks, such as object removal with physically consistent outcomes, offer a complementary diagnostic for understanding whether models capture the causal relationships between objects and their surrounding environment.

\section{Differences from Existing Benchmarks}
\label{sec:benchmark-diff}
Existing video object removal benchmarks primarily focus on spatial reconstruction or segmentation quality, and therefore do not fully evaluate causal object-environment disentanglement. In realistic scenes, removing an object requires not only filling the masked region but also accounting for secondary visual effects induced by the object, such as shadows, reflections, illumination changes, and dynamic traces. Evaluating whether these effects have been correctly removed requires temporally aligned reference videos that represent the counterfactual scene state after the object has been removed.

Traditional video segmentation datasets such as DAVIS~\cite{Perazzi_2016_CVPR} and YouTube-VOS~\cite{Xu_2018_ECCV} provide high-quality object masks and have been widely used to evaluate video inpainting methods. However, these datasets do not include aligned clean reference videos, making it difficult to determine whether an object removal method has successfully eliminated both the object and its associated environmental effects. As a result, evaluation is typically based on perceptual similarity or local reconstruction quality rather than causal correctness.

More recent datasets attempt to introduce paired references for evaluation. For example, HQVI~\cite{cho2024rgvi} provides paired videos for video inpainting evaluation. However, the dataset construction largely focuses on spatial reconstruction, and it does not explicitly model secondary object-induced effects. Consequently, it primarily evaluates the ability of a model to reconstruct missing regions rather than its ability to remove the broader physical consequences of an object’s presence.

Among existing benchmarks, ROSE-Bench~\cite{miao2025rose} is closest in spirit to our setting because it explicitly considers object-related side effects in synthetic paired videos. This represents an important step toward more physically grounded evaluation of object removal methods. ROSE-Bench constructs its dataset using game-engine-based rendering pipelines, which allow precise control over scene configuration and object interactions. Such simulation pipelines are valuable for controlled analysis, as they provide consistent geometry, lighting, and motion patterns across paired sequences.

BeyondMasks follows a complementary design philosophy aimed at increasing visual and causal diversity. First, the dataset combines synthetic paired videos with real-world paired captures, enabling evaluation under both controlled interventions and realistic visual conditions. Second, our synthetic subset is generated using modern video generation models rather than game-engine simulation, which introduces a broader range of scene appearances, motion patterns, and visual variability. Third, the dataset covers a wider variety of object-induced after-effects, including photometric effects (e.g., shadows and reflections) as well as dynamic and contact-driven traces. Fourth, the object masks intentionally exclude these secondary effects, requiring models to infer and remove them through scene reasoning rather than explicit supervision. Finally, BeyondMasks supports both mask-guided removal and instruction-driven editing, making it suitable for evaluating modern video editing and video foundation models.

Overall, the key distinction of BeyondMasks is its emphasis on \emph{causal diversity} and \emph{interventional realism}. While prior benchmarks have been valuable for studying spatial reconstruction or specific classes of side effects, our dataset is designed to better reflect the full complexity of object removal in realistic scenes, where objects interact with their environment through lighting, geometry, materials, and temporal dynamics.

\section{Additional Qualitative Results}
\label{sec:qualitative}
\begin{figure}[!h]
\centering
\includegraphics[width=0.87\linewidth]{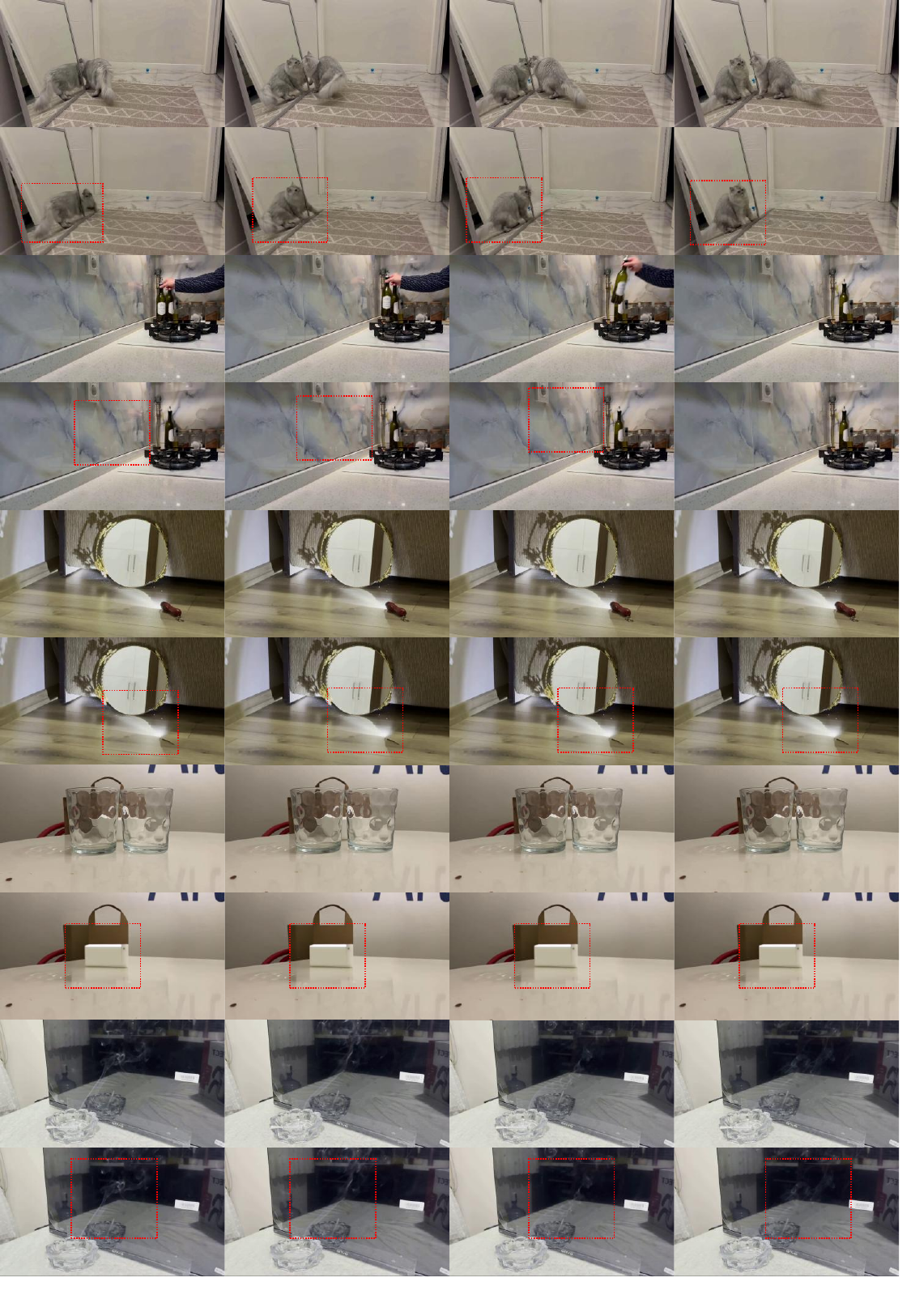}
\caption{\textbf{Failure cases in real-world videos.} For each example, the first row shows frames from the input video and the second row shows the output of a representative mask-based removal method. Although the target object is largely removed, residual object-induced effects remain visible. Red dashed boxes highlight typical artifacts such as persistent shadows, reflections in mirrors or glass surfaces, steam or smoke traces, and lighting inconsistencies that reveal incomplete removal of object-environment interactions.}
\label{fig:error_real}
\end{figure}

\begin{figure}[!h]
\centering
\includegraphics[width=0.87\linewidth]{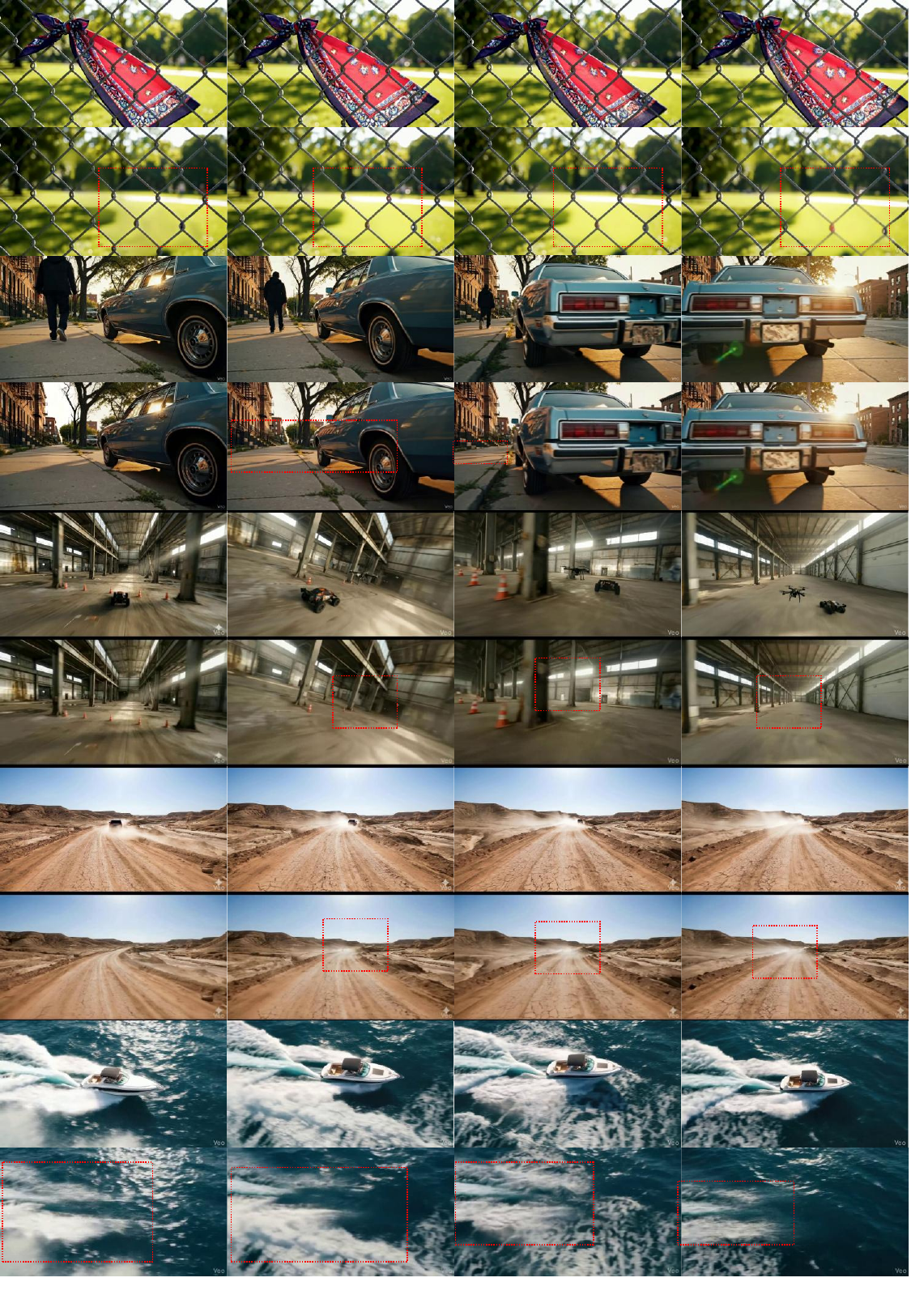}
\caption{\textbf{Failure cases in synthetic scenes.} The first row shows input frames and the second row shows the corresponding outputs after object removal. Despite successful removal of the foreground object, the edited results often retain interaction effects such as fence shadows, motion-induced dust trails, illumination artifacts, or water disturbances. Red dashed boxes indicate regions where the model reconstructs a visually plausible background but fails to remove the full causal footprint of the object.}
\label{fig:error_synthetic}
\end{figure}

We present additional qualitative examples illustrating common failure modes of current video object removal systems. While many approaches successfully remove the visible object, they often fail to eliminate the physical effects induced by the object in the surrounding environment. These effects include shadows, reflections, translucent distortions, lighting changes, and dynamic interaction traces. Such artifacts remain visible even after the object itself disappears, revealing incomplete removal of the object's causal footprint. The examples in Fig.~\ref{fig:error_real} and Fig.~\ref{fig:error_synthetic} highlight these failure patterns in both real-world and synthetic scenes.

\end{document}